\documentclass[10pt,twocolumn,letterpaper]{article}

\usepackage[pagenumbers]{cvpr} % To force page numbers, e.g. for an arXiv version
\usepackage{multirow}
\usepackage[accsupp]{axessibility}
\definecolor{cvprblue}{rgb}{0.21,0.49,0.74}
\usepackage[pagebackref,breaklinks,colorlinks,allcolors=cvprblue]{hyperref}

\def\paperID{8444} % *** Enter the Paper ID here
\def\confName{CVPR}
\def\confYear{2025}

\title{Instant Gaussian Stream: Fast and Generalizable Streaming of Dynamic Scene Reconstruction via Gaussian Splatting }

\author{Jinbo Yan\textsuperscript{1}, Rui Peng\textsuperscript{1,2}, Zhiyan Wang\textsuperscript{1}, Luyang Tang\textsuperscript{1,2},
Jiayu Yang\textsuperscript{1,2}\\Jie Liang\textsuperscript{1}, Jiahao Wu\textsuperscript{1}, Ronggang Wang\textsuperscript{1,2}\thanks{Corresponding author.}
\\
\textsuperscript{1}Guangdong Provincial Key Laboratory of Ultra High Definition Immersive Media Technology,\\ Shenzhen Graduate School, Peking University\\
\textsuperscript{2}Pengcheng Laboratory\\
{\tt\small \{yjb, ruipeng, zywang23, tly926, liangjie, wjh0616\}@stu.pku.edu.cn} 
\\
{\tt\small jiayuyang@pku.edu.cn\quad rgwang@pkusz.edu.cn}
}

\begin{document}
% \maketitle
\twocolumn[{%
\renewcommand\twocolumn[1][]{#1}%
\maketitle
\centering
\includegraphics[width=\linewidth]{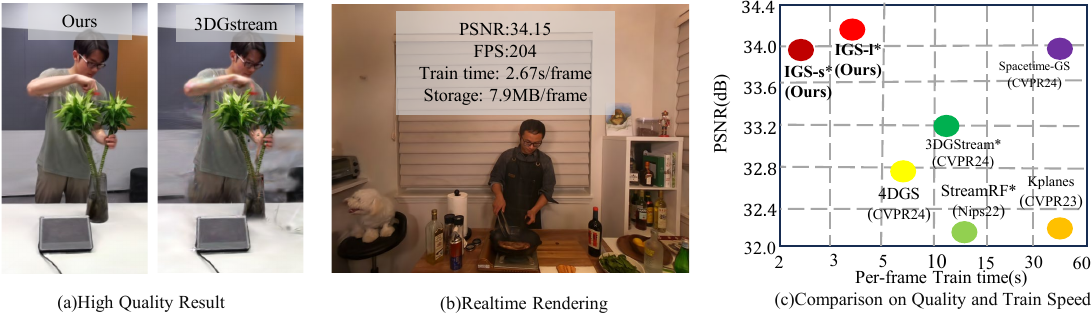}

\captionof{figure}{Performance comparison with pervious SOTA\cite{sun20243dgstream,li2023spacetime,Wu_2024_CVPR,li2022streaming,kplanes}. Our method achieves a per-frame reconstruction time of 2.67s, delivering high-quality rendering results in a streaming fashion (a)(b), with a noticeable improvement in performance (c). * denotes a streamable method.
% 我们的方法可以实现2.67s的per-frame reconstruction time,以Streaming的形式得到高质量的渲染结果(a)(b), achieving a certain improvement in performance(c). * donate a Streamable method
}}
\label{fig:teaser}
]

\begin{abstract}
% 以Streaming的方式去建立Free-Viewpoint Videos 相比于offline training methods有低时延的优势，greatly enhancing user experience.
% Streaming reconstruction of dynamic scenes is a frame-by-frame approach to capturing dynamic environments. Compared to offline training methods, streaming reconstruction offers lower latency, greatly enhancing user experience. 
% Building free-viewpoint videos in a streaming manner offers the advantage of lower latency compared to offline training methods, greatly enhancing user experience.
Building Free-Viewpoint Videos in a streaming manner offers the advantage of rapid responsiveness compared to offline training methods, greatly enhancing user experience. However, current streaming approaches face challenges of high per-frame reconstruction time (10s+) and error accumulation, limiting their broader application. In this paper, we propose Instant Gaussian Stream (IGS), a fast and generalizable streaming framework, to address these issues. First, we introduce a generalized Anchor-driven Gaussian Motion Network, which projects multi-view 2D motion features into 3D space, using anchor points to drive the motion of all Gaussians. This generalized Network generates the motion of Gaussians for each target frame in the time required for a single inference.
Second, we propose a Key-frame-guided Streaming Strategy that refines each key frame, enabling accurate reconstruction of temporally complex scenes while mitigating error accumulation. We conducted extensive in-domain and cross-domain evaluations, demonstrating that our approach can achieve streaming with a average per-frame reconstruction time of 2s+, alongside a enhancement in view synthesis quality. 

% ---

% ### Suggestions for Improvement:

% - Specify the advantage of low latency in terms of PSNR in user scenarios, as latency is typically crucial for real-time applications.
% - Consider providing a brief description of PSNR as a key metric, as “view synthesis quality” might not be immediately clear to all reviewers.
  
% This polished version should communicate your ideas clearly and professionally for an academic or reviewer audience.
% 动态场景的流式重建以frame by frame的方式来重建一个动态场景，相比于offline training的方式具有低时延的优势，极大的提高用户体验。但是目前的流式的方法面临着高时延和误差累积的问题，限制了它的进一步应用。在本篇文章中，我们提出了Instant Gaussian Streaming(IGS)来解决上述问题.首先，我们训练了一个泛化的Anchor-driven Gaussian Motion Network, 将Multi-view 的2D的Motion embedding投影到3D空间，以anchor来驱动整体Gaussian的Motion，能以一次模型inference的时间得到目标帧的Gaussian. 其次，我们提出了Key-frame guided streaming strategy,对 Key-frame进行Max points bounded的优化，能够很好的重建如物体出现消失的temporally complex的场景的同时可以消除误差累积的影响。我们进行了广泛的In-domain evaluation和Cross-domain evaluation，我们的方法可以以1.6s每帧的时延实现Streaming动态场景重建同时实现大幅的视图合成质量的提升.
\end{abstract}
\section{Introduction}
% 根据输入的Multi-view Images重建Free-Viewpoint Videos(FFV)是一个非常有价值的领域，支持许多着如VR,AR,sports broadcasting等沉浸影像的应用. 通过提供照片级真实的影像和自由视点的交互，the FFV construced from dynamic scenes 有潜力作为新一代影像媒介. To 进一步提升用户体验，流式构建FFV以frame by frame的方式去重建动态场景，相比于Offline training的方法实现了快速响应.
% However 
Reconstructing Free-Viewpoint Videos (FVV) from multi-view images is a valuable area of research, with applications spanning immersive media such as VR, AR, and sports broadcasting. By enabling interactive, photorealistic visuals, FVVs from dynamic scenes hold the potential to become a next-generation visual medium, offering experiences that go beyond traditional video formats. To enhance user experience, streaming-based FVV construction—where dynamic scenes are reconstructed frame by frame—offers a low-delay response compared to traditional offline training approaches, making it better suited for real-time, interactive applications.

With advancements in real-time rendering and high-quality view synthesis powered by 3D Gaussian Splatting (3DGS)\cite{kerbl3Dgaussians}, dynamic scene reconstruction has seen rapid progress.  
% Some offline training methods\cite{Wu_2024_CVPR,li2023spacetime,10.1145/3664647.3681463,huang2023sc,yang2023deformable3dgs,yang2023gs4d} achieve high-quality view synthesis but require complete video sequences for extensive offline training, limiting their practical application, e.g., living  streaming.
Some offline training methods\cite{Wu_2024_CVPR,li2023spacetime,10.1145/3664647.3681463,huang2023sc,yang2023deformable3dgs,yang2023gs4d} achieve high-quality view synthesis but require collecting all frames before training can begin. This limitation makes them less suitable for scenarios that demand fast response times, such as live streaming and virtual meetings.
To address these challenges, some methods\cite{sun20243dgstream,li2022streaming} adopt a streaming framework that reconstructs dynamic scenes frame by frame by modeling inter-frame differences. 
% 3DGStream, with a 12-second per-frame reconstruction time, enables online reconstruction of dynamic scenes, significantly expanding the applications of dynamic scene reconstruction.
% For example, 3DGStream achieves a per-frame latency of 12 seconds, enabling online reconstruction of dynamic scenes from video streams, thereby significantly expanding the potential applications of dynamic scene reconstruction.
However, streaming-based dynamic scene reconstruction still faces significant challenges. First, current methods typically require per-frame optimization, resulting in high per-frame latencies(10s+), which severely impact the real-time usability of these systems.
% For instance, the 12-second latency per frame in existing methods can lead to delays that hinder interactive or real-time use cases. 
Additionally, error accumulation across frames degrades the reconstruction quality of later frames, making it difficult for streaming methods to scale effectively to longer video sequences.
% 基于3DGS的实时渲染和高质量的视图合成能力，动态场景的重建也进入了快速的发展. While the offline traning methods使用规范场和变形场的解耦思路or lifting 3D Gaussian primitives to 4D space with a added temporal position, they can 实现较高的视图合成质量 but require complete video sequences
% for long time offline training, which 限制了他们的实际应用. To handle the problems, 一些方法采用Streaming的框架，通过建模帧间差异来frame by frame的重建动态场景。如3DGStream可以实现12s每帧的时延来产生FFV，enbale the online reconstruction of dynamic
% scenes in video streams, 极大的扩展了动态场景重建的应用范围。 However, Streaming的动态场景重建仍然存在挑战，首先目前的方法需要逐帧的Optimization，十多秒的逐帧时延太高。其次，误差累积会影像Streaming方法的后面帧的重建效果，which 使其难以扩展到长视频序列.

To promote the streaming framework to be more practical, we introduce Instant Gaussian Stream (IGS), a streaming approach for dynamic scene reconstruction that achieves a per-frame reconstruction time of 2s+, mitigates error accumulation, and enhances view synthesis quality. First, to tackle the issue of high per-frame reconstruction time, we developed a generalized Anchor-driven Gaussian Motion Network (AGM-Net). 
% This network employs anchor points to carry motion features that guide Gaussian transformations and enables inference to compute the motion of Gaussian primitives between frames in a single feed forward, eliminating the need for per-frame optimization. 
This network utilizes a set of key points, called anchor points, to carry motion features that guide Gaussian transformations. This design allows the inference process to compute the motion of Gaussian primitives between frames in a single feedforward pass, eliminating the need for per-frame optimization.
Second, to further improve view synthesis quality and minimize error accumulation, we propose a Key-frame-guided Streaming strategy. By establishing key-frame sequences and performing max-point-bounded refinement on key frames, our method mitigates the impact of error accumulation and enhances rendering quality in temporally complex scenes.
% 为了解决上述问题，我们提出了Instant Gaussian Streaming, a streaming method 来重建动态场景 with 每帧只需要1.6s的时延，并且缓解了误差累积的影响，提升了视图渲染的质量。First,为了解决per-frame latency 太大问题， we 构建了一个泛化的 Anchor-driven Gaussian Motion Network， 用anchor points来携带Motion feature，并驱动Gaussians的移动。which可以一次inference得到Gaussian primitives的从上一帧到当前帧的Motion。这样我们就避免了逐帧优化所来的长时延。 Second， 为了进一步提升视图合成质量以及避免误差累积带来的影响，我们提出了Key-frame based streaming 策略，通过建立key-frame 序列并对key-frame进行Max point bouned的Optimization, 我们可以缓解误差累积带来的影响，并在temporally complex的场景中取得更好的视图合成效果.
% 我们在的In-domain和cross-domain的场景下进行了广泛的验证, 实验结果证明了我们的模型的泛化性，在逐帧的时延和渲染质量上都显著由于目前的Sota方法. Our contributions are summarized below.

We conducted extensive validation in both in-domain and cross-domain scenarios, and the experimental results demonstrate the strong generalization capability of our model, with significant improvements over current state-of-the-art methods in terms of per-frame reconstruction time and rendering quality. To the best of our knowledge, this is the first approach to use a generalized method for streaming reconstruction of dynamic scenes. Our contributions are summarized below.
\begin{itemize}
    \item We propose a generalized Anchor-driven Gaussian Motion Network that captures Gaussian motion between adjacent frames with a single inference, eliminating the need for frame-by-frame optimization. 
    % 我们提出了一个泛化的Anchor-driveb Gaussian Motion Network, 以一次inference的代价得到相邻帧之间Gaussian的Motion，无需逐帧的优化。据我们所知，我们是第一个使用泛化的方法来解决动态场景的流式重建。
    \item We designed a Key-frame-guided Streaming strategy to enhance our method's capability in handling temporally complex scenes, improving overall view synthesis quality within the streaming framework and mitigating error accumulation.
    \item The evaluation results in both in-domain and cross-domain scenarios demonstrate the generalization capability of our method and its state-of-the-art performance. We achieve a 2.7 s per-frame reconstruction time for streaming, representing a significant improvement over previous methods. Additionally, we improve view synthesis quality, enabling real-time rendering at 204 FPS while maintaining comparable storage overhead.
    % 在in-domain和cross-domain上的评测结果证明了我们方法的泛化能力以及SOTA的性能。我们可以实现1.6s的时延在1352x1024 分辨下进行Streaming，相比于之前的方法有10x的提升。并且我们在视图合成质量上有大幅提升，实现实时渲染(191FPS)并且保持相近的存储开销。
\end{itemize}
\section{Related work}
\subsection{3D Reconstruction and View Synthesis}
% 新视角合成一直都是计算机视觉领域的热点问题。通过使用MLP隐式地表示场景，NeRF实现了逼真的渲染。后续的工作在vanilla NeRF的基础上提高渲染质量、减少训练视角、减轻对相机位姿的依赖、提升训练和推理速度。然而，这些工作都需要昂贵的计算代价，没有实现实时渲染。
Novel view synthesis (NVS) has always been a hot topic in the field of computer vision. By using MLP to implicitly represent the scene, Neural Radiance Fields (NeRF) \cite{nerf} achieves realistic rendering. Subsequent works have imporved NeRF to enhance rendering quality \cite{barron2021mip, barron2022mip, wang2023f2}, reduce the number of training views \cite{yang2023freenerf,wang2023sparsenerf,niemeyer2022regnerf,wu2024reconfusion}, lessen dependence on camera poses \cite{lin2021barf,truong2023sparf,chen2023local,bian2023nope}, and improve both training and inference speeds \cite{muller2022instant,fridovich2022plenoxels,  hu2023tri, barron2023zip, hedman2021snerg, reiser2021kilonerf,reiser2023merf,SunSC22, chen2022tensorf}. 
% 为了解决上述问题，3DGS使用各向异性的高斯原语表示场景并引入光栅化进行渲染，提升了速度和渲染质量。一些方法聚焦于减少内存消耗、表面重建提高几何质量、将相机位姿与高斯场联合优化，以及使用扩散模型生成高斯场。
3D Gaussian Splatting (3DGS) \cite{kerbl3Dgaussians} employs anisotropic Gaussian primitives to represent scenes and introduces rasterization-based splatting rendering algorithm, enhancing both speed and rendering quality. Some methods focus on various aspects of improving Gaussian field representations, including rendering quality\cite{lu2024scaffold,yu2024mip, ren2024octree,li20243d,zhang2024pixelgs, yang2024spec}, enhancing geometric accuracy\cite{zhang2024rade,huang20242d, Yu2024GOF}, and increasing compression efficiency, \cite{chen2025hac,yang2024spectrally,lu2024scaffold,fan2023lightgaussian}, joint optimization of camera pose and gaussian fields \cite{fan2024instantsplat,Fu_2024_CVPR,schmidt2024noposegs}, as well 3D generation \cite{zou2024triplane, tang2023dreamgaussian,tang2024lgm,LaRa}. 

\subsection{Generalizable 3D Reconstruction for Acceleration}
% 3DGS需要进行逐场景优化才能得到逼真的渲染效果。为了加速这一耗时的过程，一些工作受到泛化NeRF的启发，提出在大数据集上训练gs泛化模型，实现快速重建。pixelSplat使用an epipolar Transformer编码特征并解码为高斯属性。其他一些泛化模型使用Transformer或Multi-View Stereo构建cost volume并解码，实现了实时的渲染速度和优秀的泛化性。据我们所知，我们的工作是第一个将泛化模型用于动态流式场景的，利用其快速推理帧间运动实现对流式场景的加速。
3DGS requires per-scene optimization to achieve realistic rendering results. To accelerate this time-consuming process, some works \cite{jin2024lvsm,szymanowicz2024splatter,zheng2024gps,szymanowicz2024flash3d,li2024ggrt,zhang2024transplat}, inspired by generalizable NeRF \cite{wang2021ibrnet,yu2021pixelnerf,chen2021mvsnerf,johari2022geonerf,xu2023murf}, have proposed to train generalizable Gaussian models on large-scale datasets to enable fast reconstruction. PixelSplat \cite{charatan2024pixelsplat} utilizes an Transformer to encode features and decode them into Gaussian attributes. Other generalizable models \cite{zhang2025gs,chen2025mvsplat,liu2025mvsgaussian,fei2024pixelgaussian} utilize Transformers or Multi-View Stereo (MVS) \cite{yao2018mvsnet} techniques to construct cost volumes followed by a decoder, achieving real-time rendering speeds and excellent generalizability. To the best of our knowledge, our work is the first to apply generalizable models to dynamic streaming scenes, utilizing their rapid inference capabilities to accelerate the processing of dynamic scenes reconstruction.
% \subsection{4D的动态}
\subsection{Dynamic Scene Reconstruction and View Synthesis}
% 将静态场景重建扩展到动态场景已经有很多的尝试,一些基于NeRF的方法modeling the
% entire scene as a canonical field and a deformation field or reduce the dimensionality of the 4D space by decomposing it into a set of planar grids or hash grids.
% There have been many attempts to extend static scene reconstruction to dynamic scenes. Some methods based on NeRF model the entire scene as a canonical field and a deformation field, while others reduce the dimensionality of 4D space by decomposing it into a set of planar or hash grids. 3DGS高斯出现以来,出现了许多工作尝试把Gaussian Splatting的实时渲染的特性引入动态场景重建中. 4DGS通过一个Hexplane来建模时域上的变化,延续了规范场和变形场的思路.另一些方法lifting 3D Gaussian primitives to 4D space with a added temporal position. they can 实现较高的视图合成质量, 但是需要整个序列进行offline training,在一些需要实时交互和快速相应的场景中不适用,限制了他们的应用场景.
There have been numerous efforts to extend static scene reconstruction to dynamic scenes based on NeRF\cite{dnerf,hypernerf,devrf,neuraltraj,flowforwardnerf,kplanes,hexplane,tensor4d,im4d}.
% Some NeRF-based methods \cite{dnerf,hypernerf,devrf,neuraltraj,flowforwardnerf}model the entire scene as a canonical field and a deformation field, while others\cite{kplanes,hexplane,tensor4d,im4d} reduce the dimensionality of 4D space by decomposing it into planar or hash grids. 
Since the advent of 3D Gaussian Splatting (3DGS)\cite{kerbl3Dgaussians}, researchers have explored incorporating its real-time rendering capabilities into dynamic scene reconstruction\cite{Wu_2024_CVPR,yang2023deformable3dgs,huang2023sc,li2023spacetime,10.1145/3664647.3681463,yang2023gs4d}. 
% For instance, 4D Gaussian Splatting (4DGS)\cite{Wu_2024_CVPR} continues the approach of canonical and deformation fields. Other methods\cite{li2023spacetime,10.1145/3664647.3681463,yang2023gs4d} lift 3D Gaussian primitives into 4D space by adding a temporal position, achieving high-quality view synthesis. 

% However, these approaches rely on offline training with full video sequences, making them unsuitable for applications requiring real-time interaction and quick response, thus limiting their practical utility. 为了解决这个问题, StreamRF formulate the dynamic
% modeling problem with an incremental learning paradigm . NeRFPlayer 采用了一个. A timedependent sliding window 在feature channels从而实现Stream able. ReRF 通过DVGO 模拟a residual radiance field to enable highly compressible and streamable radiance
% field modeling. 3DGStream基于Gaussian Splatting, 通过优化一个Neural Transformation Cache来建模Gaussians的移动,进一步提升了流式动态场景建模的性能. 这些方法都取得了不错的效果,但是,都需要逐帧的优化同时带来了较长时延(目前的SOTA:10s+). 我们的方法为Stream 的动态场景建模提供了新思路,通过训练泛化的网络,可以避免逐帧的优化,实现低时延和较高的渲染质量.
However, these approaches rely on offline training with full video sequences, making them unsuitable for applications requiring real-time interaction and fast response. To address this issue, existing methods such as StreamRF\cite{li2022streaming}, NeRFPlayer\cite{song2023nerfplayer}, ReRF\cite{wang2023neural}, and 3DGStream\cite{sun20243dgstream}  reformulate the dynamic modeling problem using a Streaming method. Notably, 3DGStream\cite{sun20243dgstream}, based on Gaussian Splatting, optimizes a Neural Transformation Cache to model Gaussian movements between frames, further improving the performance. Although these methods achieve promising results, they still rely on per-frame optimization, resulting in significant delays (with current SOTA methods requiring over 10 seconds per frame\cite{sun20243dgstream}). Our approach offers a new perspective for streaming dynamic scene modeling: by training a generalized network, we eliminate the need for per-frame optimization, achieving low per-frame reconstruction time alongside high rendering quality.

% StreamRF\cite{li2022streaming} reformulates the dynamic modeling problem using an incremental learning paradigm. NeRFPlayer\cite{song2023nerfplayer} uses a time-dependent sliding window in the feature channels to enable streamable modeling. ReRF\cite{wang2023neural} simulates a residual radiance field through DVGO\cite{SunSC22} to create a highly compressible and streamable radiance field. 3DGStream\cite{sun20243dgstream}, based on Gaussian Splatting, optimizes a Neural Transformation Cache to model Gaussian movements between frames, further improving the performance. Although these methods achieve promising results, they still rely on per-frame optimization, resulting in significant delays (with current SOTA methods requiring over 10 seconds per frame\cite{sun20243dgstream}). Our approach offers a new perspective for streaming dynamic scene modeling: by training a generalized network, we eliminate the need for per-frame optimization, achieving low per-frame reconstruction time alongside high rendering quality.

% 一些基于NeRF的方法已经对建模动态场景做了很多的尝试，他们或将场景表示为

% \input{sec/preliminary}
\begin{figure*}[h]
  \centering
    % \fbox{\rule{0pt}{2.5in} \rule{0.9\linewidth}{0pt}}
    % \setlength{\abovecaptionskip}{0.cm}

  \includegraphics[width= 
   \linewidth]{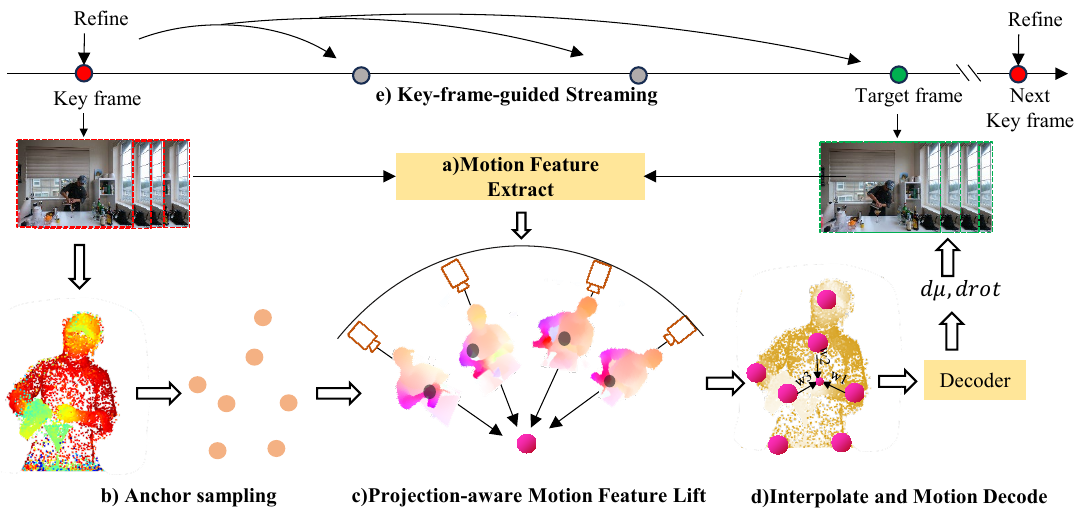}
  \caption{
  The overall pipeline of IGS. (a) Starting from the key frame and moving towards the target frame, we extract the 2D Motion Feature Map. (b) Then we sample M anchor points from the Gaussian primitives of the key frame, (c) and the anchor points are projected onto these feature maps to obtain 3D motion features through Projection-aware Motion Feature Lift. (d) Each Gaussian point interpolates its own motion feature from neighboring anchors and applies a weighted aggregation of features, which is then decoded into the motion of the Gaussian between the key frame and the target frame. (e) The entire streaming reconstruction process is guided by the Key-frame-guided Streaming strategy, where the key frame directly infers subsequent candidate frames until the next key-frame is reached, at which point max-point bounded refinement is applied to the key-frame.
  % The overall pipeline of IGS. (a) Start from Key-frame, 我们对它的Gaussian采用得到M个anchor points. (b) We can extract 一个Multi-view 的Motion Feature Maps, from Multi-view image pairs, then Project the anchor points onto the Feature Maps to get the 3D Motion Features using Projection-aware Motion Feature lifting. (c) Each Gaussian points can get 得到它自己的Motion feature通过在邻域内的Anchor插值并对特征进行加权， and then the Motion feature can decode into the Motion of the Gaussian between the key-frame and the target frame. (d) 整个流式的重建的过程被Key-frame-guided streaming策略控制, the keyframe 将会直接推理出后续的candidate frame， 知道我们到达下一个key-frame.并对key-frame施加 Max-points bounded refinement.
  }
  % \Description{ Fully described in the text.}
  \label{fig:pipeline}
  \vspace{-0.5cm}
\end{figure*}

\section{Method}

% \subsection{Pipeline}
% Given the current multiview image input, we need to deform the Gaussian primitives from the previous time step to the current one. To address this, we adopt a generalized approach that allows us to compute the displacement of the Gaussian points in a single inference, eliminating the need for frame-by-frame optimization.

% Our goal is to deform the Gaussian primitives from the previous time step to the current one, with low latency, using information from the difference between current multi-view image and 上一帧的 multi-view image . To accomplish this, we adopt a generalized approach that allows for calculating the displacement of Gaussian points in a single inference, removing the need for frame-by-frame optimization. Given the Multiview 
In this section, we begin with an overview of the pipeline in Sec.\ref{sec:overview}. 
% Then we introduce the Anchor-driven Gaussian Motion Network (AGM-Net) in Sec.\ref{sec:AGM}, which is a generalized model that drives Gaussian motion from the previous frame using anchor points. 
Then, in Sec.~\ref{sec:AGM}, we introduce the Anchor-driven Gaussian Motion Network (AGM-Net), a generalized model that drives Gaussian motion from the previous frame using anchor points, which serve as key points in the 3D scene.
Following this, we present our Key-frame-guided Streaming strategy in Sec.\ref{sec:keyframe}. Finally, in Sec.\ref{sec:training}, we outline the loss function used in our training.
\subsection{Overview}
\label{sec:overview}
Our goal is to model dynamic scenes in a streaming manner with minimal per-frame reconstruction time.
To achieve this, we adopt a generalized AGM-Net that extracts 3D motion features from the scene using anchor points and drives the motion of Gaussian primitives between frames in a single inference step.
% To achieve this, we adopt a generalized AGM-Net that extracts 3D motion features from the scene using anchor points to drive Gaussian motion between frames in a single inference step.
And we propose a key-frame-guided Streaming strategy to further improve view synthesis quality and handle temporally complex scenes while addressing error accumulation in streaming reconstruction. The overall pipeline is illustrated in Fig. \ref{fig:pipeline}.
\subsection{Preliminary}

% \subsection{3DGS}
% gaussian splatting将静态场景表示为一组各向异性的3d gaussians，每个像素的颜色通过基于点的alpha混合渲染来获得，从而实现高保真度的实时新视角合成。
Gaussian splatting\cite{kerbl3Dgaussians} represents static scenes as a collection of anisotropic 3D Gaussians. The color of each pixel is obtained through point-based alpha blending rendering, enabling high-fidelity real-time novel view synthesis.

% 具体而言，每个高斯原语包含中心μ、3D协方差矩阵Σ、不透明度α和使用n阶球谐系数表示的颜色c，可表示为
Specifically, each Gaussian primitive $\mathcal{G}_i$ is parameterized by a center $\mu \in \mathbb{R}^3$, 3D covariance matrix $\Sigma \in \mathbb{R}^{3 \times 3}$, opacity $\alpha \in \mathbb{R}$, and color $c \in \mathbb{R}^{3(n+1)^2}$
% which is represented by n-degree SH coefficients
:
\begin{equation}
  \mathcal{G}(x) = e^{-\frac{1}{2}(x-\mu)^T\Sigma^{-1}(x-\mu)}
  \label{3dgs_define}
\end{equation}

% 为了保持协方差矩阵的物理意义，它必须是半正定的。因此，将协方差矩阵分解为一个缩放矩阵S和一个旋转矩阵R，其中S和R分别使用一个缩放向量和一个四元数旋转向量来表示。
% To maintain the physical meaning of the covariance matrix, it must be positive semi-definite.Therefore, the covariance matrix $\Sigma$ can be decomposed into a scaling matrix $S$ and a rotation matrix $R$:
% \begin{equation}
%   \Sigma = RSS^T R^T
%   \label{3dgs_decomposed}
% \end{equation}

% 渲染时首先将3D gaussian投影到2D空间。给定视角变换W，可以计算得到2D协方差矩阵，其中J是the Jacobian of the affine approximation of the projective transformation.随后基于深度对覆盖一个像素的高斯进行排序，使用基于点的alpha混合渲染得到像素的颜色。
During rendering, the 3D Gaussian is first projected onto 2D space. 
% Given a view transformation matrix $W$, the 2D covariance matrix $\Sigma'$ can be computed as follows:
% \begin{equation}
%   \Sigma' = JW\Sigma W^T J^T
%   \label{3dgs_to2D}
% \end{equation}
% where $J$ is the Jacobian of the affine approximation of the projective transformation.
Subsequently, the Gaussians covering a pixel are sorted based on depth. The color of the pixel $\mathbf{c}$ is obtained using point-based alpha blending rendering:
\begin{equation}
  \mathbf{c} = \sum_{i=1}^n c_i \alpha_i^{\prime} \prod_{j=1}^{i-1}(1-\alpha_i^{\prime})
  \label{3dgs_render}
\end{equation}
Here, ${\alpha^{\prime}}$ represents the opacity after projection onto the 2D space.
% 此外，3DGS还引入了自适应密度控制。对view-space positional gradients大的高斯进行克隆和分裂并周期性地剪枝在世界空间中非常大的和不透明度过小的高斯分布以提高渲染质量。
% Additionally, gaussian splatting introduces adaptive density control, which clone and split Gaussians that exhibit large view space positional gradients, while periodically pruning Gaussians that are excessively large in world space or have low opacity to enhance rendering quality.

\subsection{Anchor-driven Gaussian Motion Network}
\label{sec:AGM}

\noindent\textbf{Motion Feature Maps:}
%这里应该重点讲的是怎么搞backbone，并且做了depth-view conditon 的finetune
 Given multi-view images of current frames $\mathbf{I}^{'}=(I_1^{'},...,I_V^{'})$ with camera parameters
 % $\mathbf{\pi}= (\pi_1, . . . , \pi_V)$
 , We can first construct a multi-view image pair, which contains the current frame and the previous frame $\mathbf{I}$ from corresponding viewpoints. 
Then, we use a optical flow model to obtain the intermediate flow embeddings. Next, a modulation layer\cite{Peebles2022DiT,LaRa} is applied to inject the viewpoint and depth information into the embeddings, ultimately resulting in 2D motion feature maps $\mathbf{F} \in \mathbb{R}^{V\times C \times H \times W}$.
\noindent\textbf{Anchor Sampling:} 
% To deform the Gaussian primitives $\mathcal{G}$ from previous frame, we need to calculate motion of each Gaussian. but 会存在计算复杂度高，显存占用大 due to 高斯点数量过多, if 我们直接对每个Gaussian进行计算。 为此we employ an anchor-point-based approach to represent the motion features of the entire scene in 3D space. the Anchor-Driven approach supports batch processing during training, reducing computational overhead,的同时能保持Gaussian primitives的几何信息.Specifically, we use farthest point sampling(FPS) to sample M anchor points from the $N$ Gaussian primitives:
To deform the Gaussian primitives \(\mathcal{G}\) from the previous frame, we need to compute the motion of each Gaussian. However, directly computing the motion for each Gaussian is computationally expensive and memory-intensive due to the large number of Gaussian points. To address this, we employ an anchor-point-based approach to represent the motion features of the entire scene in 3D space. The anchor-driven approach supports batch processing during training, reducing computational overhead while preserving the geometric information of the Gaussian primitives. Specifically, we use \textbf{Farthest Point Sampling} (\textbf{FPS}) to sample $M$ anchor points from the $N$ Gaussian primitives:

% Furthermore, by performing Projection-aware 3D Motion Feature lift, the anchor-based representation allows us to obtain higher-resolution motion features from the 2D motion feature maps, enabling each Gaussian primitive to deform more accurately.

% we employ an anchor-point-based approach to represent the motion features of the entire scene in 3D space. These motion features are computed using our Projection-aware 3D Motion Feature Lift module. 
% Compared to calculating motion features for each Gaussian primitive individually, the Anchor-Driven approach supports batch processing during training, reducing computational overhead.
% Compared to calculating motion features for each Gaussian primitive individually,采用基于Anchor Driven的方法能在训练时支持批处理，reduces computational overhead.
% our anchor-based method not only ensures stronger local consistency but also reduces computational overhead. 
% Furthermore, by performing Projection-aware 3D Motion Feature lift, the anchor-based representation allows us to obtain higher-resolution motion features from the 2D motion feature maps, enabling each Gaussian primitive to deform more accurately.

% Specifically, we use farthest point sampling(FPS) to sample M anchor points from the $N$ Gaussian primitives:
\begin{equation} 
\vspace{-0.1cm}
\mathcal{C} = \mathbf{FPS}(\{\mu_i\}_{i \in N})
\end{equation}
where $\mathcal{C}  \in \mathbb{R}^{M\times 3}$ represents the sampled anchor points with $M$ set to 8192 in our experiments, and $\mu_i$ denotes the position of  $\mathcal{G}_i$

\noindent\textbf{Projection-aware 3D Motion Feature Lift:}
We adopt a projection-aware approach to lift multiview 2D motion features into 3D space. Specifically, we project sampled anchor points onto each motion feature map based on the camera poses, obtaining high-resolution motion features:
\begin{equation}
    f_{i}=\frac{1}{V}\sum_{j \in V}\Psi( \Pi_j(\mathcal{C}_i), F_j)
\end{equation}
where \( \Pi_j(\mathcal{C}_i) \) represents the projection of \( \mathcal{C}_i \) onto the image plane of \( F_j \) using the camera parameters of \( F_j \), and \( \Psi \) denotes bilinear interpolation.
By projection, each anchor point can accurately obtain its feature $f_{i}\in \mathbb{R}^{C}$ from the multi-view feature map, effectively lifting the 2D motion map into 3D space.

We then use these features $\{f_i\}_{i\in M}$, stored at each anchor point, as input to a Transformer block using self-attention to further capture motion information within the 3D scene. 
\begin{equation}
    \{z_i: z_i \in \mathbb{R}^{C} \}_{i\in M} = \mathbf{Transformer}(\{f_i\}_{i\in M})
\end{equation}
% In this way, we effectively lift the 2D motion map into 3D space.

The output of the Transformer block $\{z_i\}_{i\in M}$ represents the final 3D motion features we obtain. Now, we can use these 3D motion features to represent the motion information of an anchor and its neighborhood, and drive the motion of the neighboring Gaussian points based on these motion features.

\noindent\textbf{Interpolate and Motion Decode:} Using the 3D motion features stored at anchor points, we can assign each Gaussian point a motion feature by interpolating from its K nearest anchors in the neighborhood:
\begin{equation}
    z_i = \frac{\sum_{k \in \mathcal{N}(i)} e^{-d_k}z_{k}}{\sum_{k \in \mathcal{N}(i)}e^{-d_k}}
    \label{interpolate}
\end{equation}
where $\mathcal{N}(i)$ represents the set of neighboring anchor points of Gaussian point $\mathcal{G}_i$, and $d_k$ represents the Euclidean distance from Gaussian point ${G}_i$ to anchor $\mathcal{C}_k$.
Then we can use a Linear head to decode the Motion feature to the movement of a Gaussian primitive:
\begin{equation}
    d\mu_i, d rot_i = \mathbf{Linear}(z_i)
\end{equation}
here, we use the deformation of the Gaussian's position $d\mu_i $, and the deformation of the rotation $drot_i$, to represent the movement of a Gaussian primitive. The new position and rotation of the Gaussian are as follows:
\begin{equation}
\mu_i^{'} = \mu_i + d\mu_i,
\end{equation}
\begin{equation}
rot_i^{'} = norm(rot_i) \times norm(drot_i).
\end{equation}
here $'$ refers to the new attributes. $norm$  denotes to quaternion normalization and $\times$ represents quaternion multiplication, as used in previous work\cite{sun20243dgstream}.

\subsection{Key-frame-guided Streaming}
\label{sec:keyframe}

Using AGM-Net, we can transition Gaussian primitives from the previous frame to the current frame within a single forward inference pass. However, this process only adjusts the position and rotation of Gaussian primitives, making it effective for capturing rigid motion but inadequate for accurately representing non-rigid motion. Furthermore, the number of Gaussian points remains constant, limiting its capacity to model temporally dynamic scenes where objects may appear or disappear. These limitations result in challenges in capturing scene dynamics and can lead to error accumulation across frames.

% To more accurately model object changes and mitigate error accumulation, we propose a Key-frame-guided Streaming strategy that selects key frames as the initial state for deforming Gaussians in subsequent frames. Additionally, we incorporate a Max-points bounded refinement strategy, which allows efficient key-frame reconstruction without redundant points and prevents iterative increases in point count across frames. This approach also mitigates over  fitting in sparse viewpoint scenes by effectively managing the point density.
To better model object changes and reduce error accumulation, we propose a Key-frame-guided Streaming strategy that uses key frames as the initial state for deforming Gaussians in subsequent frames. We also introduce a Max points bounded refinement strategy, enabling efficient key frame reconstruction without redundant points and preventing point count growth across frames. This approach helps avoid overfitting in sparse-viewpoint scenes by effectively managing point density.

\noindent\textbf{Key-frame-guided strategy:} Starting from frame $0$, we designate a key frame every $w$ frames, forming a key-frame sequence$\{K_0, K_w, ..., K_{nw}\}$ . The remaining frames serve as candidate frames. During streaming reconstruction, for example, beginning with a key frame $K_{iw}$, we deform the Gaussians forward across successive candidate frames using AGM-Net until reaching the next key frame $K_{(i+1)w}$ . At this point, we refine the  deformed Gaussians of key frame $K_{(i+1)w}$. Then, we continue deforming from key frame $K_{(i+1)w}$ to process subsequent frames.

This key-frame-guided strategy offers several advantages. First, when AGM-Net is applied to candidate frames, it is always start from the most recent key frame, preventing error propagation across candidate frames between key frames and eliminating cumulative error. Second, candidate frames do not require optimization-based refinement, as their Gaussians are generated through a single model inference with AGM-Net, ensuring low per-frame reconstrution time. Additionally, we can batch process up to $w$ frames following each key frame, which further accelerates our pipeline.
% 从第0帧开始我们每隔w帧会设立一个key frame，形成一个key frame序列{K0,K1,..,KW}， 其余的帧作为候选frame. 当进行streaming重建时，例如从一个key-frame Ki开始，我们根据Anchor based Gaussian Deforming 向后续的帧进行deform,直到我们得到了下一个key-frame K_i+1的Gaussians, then we will use Max points bounded key-frame Optimization对 key_frame k_i+1 的gaussians进行refine. Then we will start from key-frame K_i+1去进行后续的deform.

% This key-frame-guided strategy 有以下的优点, 首先，.  在使用Anchor based Gaussian Deforming得到某候选帧的Gaussian时，是以最近的上一个key-frame作为的基准，所以误差不会在两个关键帧之间的候选帧之间传播，从而消除了误差累积. 其次,在候选帧无需进行基于Optimization的refine, 他们的Gaussians都是的using our Anchor based Gaussian Deforming 通过一次模型推理得到， 所以保证了低时延.并且，我们可以对一个key-frame后的w个帧进行批处理，这可以进一步提升我们的速度.

% 这个处理可以消除误差累积
% 这个处理可以batch 处理，进一步可以加速
\noindent\textbf{Max points bounded Key-frame Refinement:}
During the refinement of each key frame, we optimize all parameters of the Gaussians and support cloning, splitting, and filtering, which is same to 3DGS\cite{kerbl3Dgaussians}. This approach allows us to handle object deformations as well as the appearance and disappearance of objects in temporally complex scenes, effectively preventing error accumulation from key frame to subsequent frames. However, this optimization strategy can lead to a gradual increase in Gaussian primitives at each key frame, which not only raises computational complexity and storage requirements but also risks overfitting in sparse viewpoints, particularly in dynamic scenes where viewpoints are generally limited.

To address this, we adopt a Max Points Bounded Refine method. When densifying Gaussian points, we control the number of Gaussians allowed to densify by adjusting each point’s gradient, ensuring that the total number of points does not exceed a predefined maximum.

\subsection{Loss Function}
\label{sec:training}

Our training process consists of two parts: offline training the generalized AGM-Net and performing online training for the key frames. The generalized AGM-Net only needs to be trained once, and it can generalize to multiple scenes. We train the AGM-Net across scenes using gradient descent, relying solely on a view synthesis loss between our predicted views and the ground truth views, which includes an $\mathcal{L}_1$ term and an $\mathcal{L}_{D-SSIM}$ term and can be formulated as:
\begin{equation}
    \mathcal{L} = (1-\lambda)\mathcal{L}_1 + \lambda\mathcal{L}_{D-SSIM}
    \label{eq:loss}
\end{equation}
When performing online training on the Gaussians in key frames, we use the same loss function as in Eq.\ref{eq:loss}. However, this time, we optimize the attributes of the Gaussian primitives rather than the parameters of the neural network.
% 我们的训练过程包含训练AGM-Net和对Key-frame的在线训练两部分. We train our AGM-Net across scenes via gradient descent, and only use 视图合成损失 between 我们预测出来的视图和GT视图,which includes a L1 term and a Lssim term, and can be formulated as:
% eq2
% 在对Key-frame 的Gaussians进行在线训练时，我们的Loss Function采用与eq2相同的形式，不过这次我们要优化的不是Neural Network的参数而是the attributes of Gaussian Primitives.

% \input{sec/stream}
\section{Implementation details}
% 在本章中，我们首先介绍了我们采用的数据集以及对于训练数据的划分和预处理 in sec 3.1。接着我们详细介绍了AGM Network的配置 in Sec3.2 以及训练时的超参数 in Sec 3.3。最后我们介绍了在streaming inference时的详细设置
In this Section, we first introduce the datasets we used, along with the partitioning and preprocessing of training data, in Sec. \ref{sec:dataset}. Next, we provide a detailed explanation of the configuration of the AGM network and the training hyperparameters in Sec. \ref{sec:network}. Finally, we describe the detailed setup for streaming in Sec. \ref{sec: streaming detail}.
\subsection{Datasets}
\label{sec:dataset}
\noindent\textbf{The Neural 3D Video Datasets (N3DV)}\cite{li2022neural} includes 6 dynamic scenes recorded using a multi-view setup featuring 21 cameras, with a resolution of 2704×2028. Each multi-view video comprises 300 frames.

\noindent\textbf{Meeting Room Datasets}\cite{li2022streaming} includes 3 dynamic scenes recorded with 13 cameras at a resolution of 1280 × 720. Each multi-view video also contains 300 frames.

\noindent\textbf{Dataset Preparation:} We split four sequences from the N3DV dataset into the training set, with the remaining two sequences, $\{cut\ roasted\ beef, sear\ steak\}$, used as the test set. For the training set, we constructed 3D Gaussians for all frames in the four training sequences, totaling 1200 frames, which required 192 GPU hours. For each frame's 3D Gaussian, we performed motions forward and backward for five frames, creating 12,000 pairs for training. 
% To evaluate our model's generalization ability, 
% we fine-tune it on the $discussion$ scene from the Meeting Room dataset for 5 epochs, accounting for variations in scale and camera parameters across different datasets, before conducting cross-domain evaluation. 
For testing, we selected one viewpoint for evaluation for both datasets, consistent with previous methods.
% To assess our model’s generalization ability, we fine-tune it on the $discussion$ scene from the Meeting Room dataset for 5 epochs, a necessary step given the variations in scale and camera parameters across different datasetsm, then do cross-domain evaluation.
% do the cross-domain evaluation, fine-tuned it on the $discussion$ scene from the Meeting Room dataset for 5 epochs, a necessary step given the variations in scale and camera parameters across different datasets. 
% This generated 300 Gaussian representations, forming 3000 training pairs. 
% We then evaluated the model on the remaining two Meeting Room sequences, $\{vrhead, trimming\}$. For testing, we selected one viewpoint for evaluation, consistent with previous methods.
% \noindent\textbf{Datasets Preparation:}
% 我们将N3DV中的4个序列划分为训练集,剩余的两个序列{cut roasted beef, sear steak}作为测试集. 在构建训练集时，我们对训练集中的四个序列的所有帧，共计1200个场景，逐个构建了3D Gaussian。接着对于每一帧的3D Gaussian，我们可以让其向前后5帧做deform，So一共有12000个Pair可以用来做训练. 由于不同数据集之间的尺度和相机参数的差异，to assess our model’s 泛化能力，我们选取了Meeting Room上的 discussion 场景做finetune,构建了300个Gaussian，形成了3000个训练Pair. 接着在剩余的Meeting Room的{VRhead, trimming}两个序列上做测试. 在做测试时，我们选取一个视角作为评测，与之前的方法相同.
\subsection{AGM Network}
\label{sec:network}
We use GM-Flow\cite{xu2022gmflow} to extract optical flow embeddings and add a Swin-Transformer\cite{liu2021Swin} block for fine-tuning while keeping the other parameters of GM-Flow fixed. Our AGM model accepts an arbitrary number of input views. To balance computational complexity and performance, we use \( V=4 \) views, each producing a motion map with \( C=128 \) channels and a resolution of 128 x 128. We sample \( M=8192 \) anchor points from Gaussian Points, which sufficiently captures dynamic details. The Transformer block in 3D motion feature lift module comprises 4 layers, yielding a 3D motion feature with \( C=128 \) channels. For rendering, we adopt a variant of Gaussian Splatting Rasterization from Rade-GS\cite{zhang2024rade} to obtain more accurate depth maps and geometry.
% We 使用GM-Flow 来提取光流embedding, 并加了一个Swin-Transformer block 来做fine tune 并且fix GM-Flow中的其他参数. AGM模型可以接收任意视角个数的输入，and 考虑计算复杂度和效果的权衡，我们选取了V=4,每个视角下得到C=128, resolution of 256 x 256的Motion Map. 我们对Gaussian Points 采样得到M=8192个anchor，which可以充分的表达动态的细节. We use the PointRasterization to do the point projection and the transformer block in 3D Motion feature lifting has 4 layers, resulting in C=128的3D Motion feature. 我们在渲染时选择了from Rade-GS的Gaussian Splatting Rasterization 的变体，从而可以得到更准确的深度图.
% \subsection{Training}
% \label{sec:train detail}

During training, we randomly select 4 views as input and use 8 views for supervision. Training is conducted on four A100 GPUs with 40GB of memory each, running for a total of 15 epochs with a batch size of 16. 
% Fine-tuning on the Meeting Room dataset takes 5 epochs with a batch size of 32.
The parameter \( \gamma \) in Eq. \ref{eq:loss} is set to 0.2. We use the Adam optimizer with a weight decay of 0.05, and \(\beta\) values of (0.9, 0.95). The learning rate is set to \(4 \times 10^{-4}\) for training on the N3DV dataset.
% and \(4 \times 10^{-5}\) for fine-tuning on the Meeting Room dataset.

\subsection{Streaming Inference}
\label{sec: streaming detail}
% We set $w=5$ for constructing keyframe sequences, yielding a total of 60 keyframes in a 300-frame video. Additionally, we conduct an ablation study to analyze the impact of varying $w$ sizes. 
We set \( w=5 \) to construct key frame sequences, resulting in 60 keyframes from a 300-frame video, and conduct an ablation study to assess the impact of different \( w \) values in Sec. \ref{sec:ablation}.
% The maximum number of points is configured at 150,000 for the N3DV dataset and 40,000 for the Meeting Room dataset. 
We designed two versions for keyframe optimization: a smaller version \textbf{IGS}-$s$ (Ours-s) with 50 iterations refinement for Key frames, providing lower per-frame latency, and a larger version \textbf{IGS}-$l$ (Ours-l) with 100 iterations, which achieves higher reconstruction quality. In both versions, densification and pruning are performed every 20 iterations. 
% For further details on learning rate parameters, please refer to the supplementary material.
For the test sequences, we construct the Gaussians for the 0th frame using the compression method provided by Lightgaussian\cite{fan2023lightgaussian}, which reduces storage usage and mitigates overfitting due to sparse viewpoints. We employ 6,000 iterations for training the first frame of the N3DV dataset, compressing the number of Gaussians at 5,000 iterations. For the Meeting Room dataset, we train the Gaussians of the first frame using 15,000 iterations, compressing the number of Gaussians at 7,000 iterations. For more details, please refer to the Supp..
% For the test sequences, we build the Gaussians for the 0th frame 用了LightGaussian提供的压缩方法，which 可以降低Storage usage并且mitigate overfitting due to sparse viewpoints. We use 6,000 iterations for training first frame of the N3DV dataset and compress the number of Gaussians at 5000 iterations. For the Meeting Room dataset, we train the Gaussians of the first frame using 15,000 iterations, compressing the number of Gaussians at 7,000 iterations.
% \subsection{Training}
% 在训练时，我们在所有视角中任选4个作为输入，然后选取8个视角作为监督。。
% 我们的训练在4张40G的A100上进行，训练共计了15个epoch,with a total batchsiaze of 16. And finetune on the Meetingroom datasets has 5 epochs，　with　a total　batchsize of 64, 用时大约1h. The gamga1 in eq12 is 0.2. 我们使用Adam Optimizer with the learning rate of 4×10−4
% , weight decay of 0.05, and betas of (0.9, 0.95)。 And learning rate is 4e-4 for training on N3DV datasets and 4e-5 for futuning on the Meeting Room dataset.
% \subsection{Streaming inference}
% We use w=5 when 建立Key frame sequences, with a total 60 Key frames in a 300 frame video. And we set the max points for the N3DV as 300000, and 40000 for the Meeting Room datasets. For the test sequences, we 预先build the Gaussians for 第0帧 using 5000 iterations for N3DV dataset. For the Meeting Room dataset, we train the first frame's Gaussian using 10000 iterations with compress the Number of Gaussians at 7000 iterations using LightGaussian, 这会缓解由于视角稀疏带来的过拟合问题.
\section{Experiments}
% \subsection{Baselines and Metrics}

% 我们与当前的Sota的动态场景重建的方法进行实验比较。他们可以主要分为两类，(1)离线训练的动态场景重建方法，包括 4DGS, Saro-GS, Spacetime-GS.(2) 在线训练的Streaming的方法, 包括3DGStream, StreamRF. 4DGS, Spacetime-GS, Saro-GS均是基于Gaussian Splatting, 用一组Gaussian Primitives来表示整个动态场景. 3DGStream和StreamRF采用在线训练的方式，逐帧的来优化.其中3DGStream通过逐帧的优化Neural Transform Cache来建模高斯点在逐帧之间的移动，建立了一个3DGS-based pipeline for Free-Viewpoint Video streaming，enabling the on-the-fly and high-quality reconstruction of dynamic scenes.

\subsection{Metrics and Baselines}
\label{sec: metrics}
% Following previous works, We 测试and report the PSNR, Storage, Training Time, FPS来与之前的方法进行比较. 上述的方法都是在.我们还额外报告了Per-Frame Latency, 记录on-the-fly中重建每一帧所需要的时间，来评价方法在Streaming中的时延，which is important for 实际应用和用户体验.

\noindent\textbf{Baselines:} We compare our approach to current state-of-the-art methods for dynamic scene reconstruction, covering both offline and online training methods. Offline methods\cite{kplanes,Wu_2024_CVPR,li2023spacetime,10.1145/3664647.3681463,yang2023gs4d} rely on a set of Gaussian primitives or Hex-planes to represent entire dynamic scenes. Online training methods\cite{sun20243dgstream,li2022streaming} employ per-frame optimization to support streaming reconstruction. Specifically, 3DGStream\cite{sun20243dgstream} models the movement of Gaussian points across frames by optimizing a Neural Transform Cache, creating a 3DGS-based pipeline for free-viewpoint video streaming that enables high-quality, real-time reconstruction of dynamic scenes. 
% And the train 

% which can be broadly categorized into two types: (1) offline-trained dynamic scene reconstruction methods, including 4DGS, Saro-GS, and Spacetime-GS; and (2) online-trained streaming methods, such as 3DGStream and StreamRF. The offline methods—4DGS, Spacetime-GS, and Saro-GS—all rely on Gaussian Splatting, using a set of Gaussian primitives to represent the entire dynamic scene. The streaming methods, 3DGStream and StreamRF, use online training to incrementally optimize frame-by-frame. Specifically, 3DGStream models the frame-by-frame movement of Gaussian points by optimizing a Neural Transform Cache, creating a 3DGS-based pipeline for free-viewpoint video streaming that enables high-quality, real-time reconstruction of dynamic scenes.
\noindent\textbf{Metrics:}
Following prior work, we evaluate and report \textbf{PSNR}, \textbf{Storage} usage, \textbf{Train} time, and \textbf{Render} Speed to compare with previous methods.
% Additionally, we introduce Per-Frame \textbf{ Latency}, which records the time required to reconstruct each frame on-the-fly, offering insight into the method’s latency in streaming scenarios—a critical factor for practical applications and user experience. 
All metrics are averaged over the full 300-frame sequence, including frame 0.
% Per-frame latency is averaged over the remaining 299 frames obtained through streaming.
% All other metrics, including PSNR, Storage, Train time, and rendering speed, are averaged over the full 300-frame sequence, including frame 0. The per-frame latency, however, is averaged over the remaining 299 frames obtained through streaming.
\subsection{Comparisons}
% 我们在N3DV上的两个测试序列上来验证我们模型的in-domn 泛化能力
% We present our in-domain evaluation on N3DV的两个测试序列，and 评测结果如表2. 为了公平的比较我们的方法的性能，我们测试了3DGStream使用与我们的第0frame的Gaussians的性能，同时使用与我们相同的variant of Gaussian Splatting Rasterization，结果如表中的 *. 相比于 3DGStream和StreamRF, 我们的方法在Per-frame 时延上有着近10x的提升，可以做到每帧平均1.6s的时延，同时保持着相近的Render Speed和Storage Usage. 我们的方法also achieves a certain level of enhancement in rendering quality.
% 相比于Offline training的方法，我们的方法可以实现低时延的Streaming，并且质量和所需的训练时间由于对于的SOTA的方法。 So,in summary, 我们的方法凭借泛化的AGM-Net和Key-frame-guided streaming策略，可以以1.6s的时延进行3DGS的 Streaming而无需逐帧优化，并且实现SOTA的渲染质量，with no 在渲染速度和Storage上的衰减。
\noindent\textbf{In-domain evaluation:} We present our in-domain evaluation on two test sequences from the N3DV dataset, with results shown in Tab. \ref{tab:n3d}. For a fair comparison of performance, we tested 3DGStream using the same Gaussians from the 0th frame and applied the same variant of Gaussian Splatting Rasterization as used in our approach (denoted with $ \dagger$ in the table). Compared to 3DGStream and StreamRF, our method achieves a 6x reduction in train time, with an average delay of 2.67 seconds per frame, while maintaining comparable rendering speed and storage usage. Our approach also achieves enhanced rendering quality. Compared to offline training methods, our approach provides low-delay streaming capabilities while achieving state-of-the-art rendering quality and reducing training time. 
A qualitative comparison of rendering quality can be seen in Fig. \ref{fig:n3dv_ qualitative}. It is evident that our method outperforms others in rendering details, such as the transition between the knife and fork, and in modeling complex dynamic scenes, like the moving hand and the shifting reflection on the wall.
% A qualitative comparison of rendering quality can be seen in Fig. 5, 可以看出我们的方法在渲染细节上(刀和叉子交接的地方)和建模复杂运动场景上(运动中的手和墙上的倒影)更有优势
% In summary, with the generalized AGM-Net and the Key-frame-guided streaming strategy, our method enables streaming-based modeling of dynamic scenes with a per-frame reconstruction time of just 2.67 seconds per frame, eliminating the need for per-frame optimization. This is achieved without any decline in rendering speed or storage efficiency, while maintaining state-of-the-art rendering quality.

\begin{table}
\centering
\caption{Comparison on the N3DV dataset, with results measured at a resolution of 1352 x 1014. $ \dagger$ indicates that the evaluation was performed using the official code in the same experimental environment as ours, including the same initial point cloud. Highlights denote the \textbf{best} and \underline{second best} results.}
% \vspace{-10pt}
\resizebox{0.45\textwidth}{!}{\begin{tabular}{lccccc}
    \toprule
\multirow{2}{*}{Method}                            & PSNR$\uparrow$                       & Train 
 $\downarrow$ & Render$\uparrow$    & Storage$\downarrow$  \\  
 &(dB)&(s)&(FPS)&(MB)\\   \midrule
 Offline training\\ \midrule
Kplanes\cite{kplanes}       & 32.17      & 48  &0.15  &1.0        \\
Realtime-4DGS\cite{yang2023gs4d}   & 33.68   & -  &114      &-        \\
4DGS\cite{Wu_2024_CVPR}    & 32.70      & 7.8  &30      &\textbf{0.3}        \\
Spacetime-GS\cite{li2023spacetime}     & 33.71     & 48     & 140   & \underline{0.7}              \\
Saro-GS\cite{10.1145/3664647.3681463}    & \underline{33.90}    & -  & 40    & 1.0         \\
\midrule
 Online training\\ \midrule
StreamRF\cite{li2022streaming}  & 32.09  & 15  & 8.3     & 31.4              \\
3DGStream\cite{sun20243dgstream} & 33.11  & 12     & \textbf{215}      & 7.8                 \\
3DGStream\cite{sun20243dgstream}$\dagger$ & 32.75  & 16.93 &\underline{204}   & 7.69                  \\
Ours-s          & 33.89 & \textbf{2.67} &\underline{204} &   7.90                \\   
Ours-l       & \textbf{34.15}  & \underline{3.35} &\underline{204}&   7.90                  \\   
\bottomrule
\end{tabular}}
\label{tab:n3d}
% \vspace{5pt}
\vspace{-0.6cm}
\end{table}

We also conducted a PSNR trend comparison with 3DGStream to verify the effectiveness of our method in mitigating error accumulation. The comparison results and the smoothed trends are shown in Fig. 2. As seen, our rendering quality does not degrade with increasing frame number, while 3DGStream suffers from error accumulation, with a noticeable decline in quality as the frame number increases. This confirms the effectiveness of our approach in addressing error accumulation. However, it is also apparent that our method exhibits more fluctuation in per-frame PSNR. This is because 3DGStream assumes small inter-frame motion, leading to smaller adjustments and smoother differences between frames.
\begin{figure}[h]
  \centering
    % \fbox{\rule{0pt}{2.5in} \rule{0.9\linewidth}{0pt}}
    % \setlength{\abovecaptionskip}{0.cm}

  \includegraphics[width= 
   0.8\linewidth]{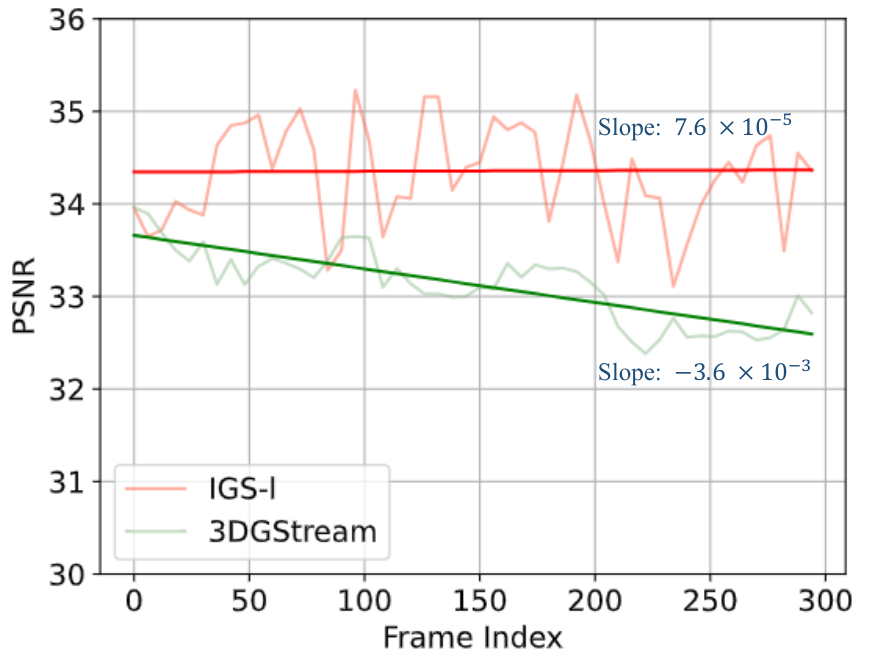}
     \vspace{-0.5cm}

  \caption{The PSNR trend comparison on the sear steak .}
  % \Description{ Fully described in the text.}
  \label{fig:perframe}
  % \vspace{-0.5cm}
\end{figure}

\begin{figure}[h]
  \centering
    % \fbox{\rule{0pt}{2.5in} \rule{0.9\linewidth}{0pt}}
    % \setlength{\abovecaptionskip}{0.cm}

  \includegraphics[width= 
   0.9\linewidth]{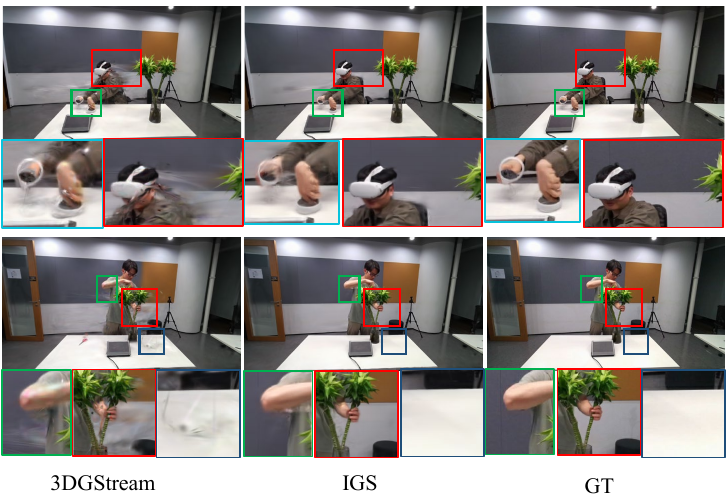}
  \caption{Qualitative comparison from the Meeting Room dataset.}
  % \Description{ Fully described in the text.}
  \label{fig:meeting}
  \vspace{-0.5cm}
\end{figure}

\begin{figure*}[h]
  \centering
    % \fbox{\rule{0pt}{2.5in} \rule{0.9\linewidth}{0pt}}
    % \setlength{\abovecaptionskip}{0.cm}

  \includegraphics[width= 
   % \linewidth]{pic/n3d_comprasion_downsize.eps}
      0.9\linewidth]{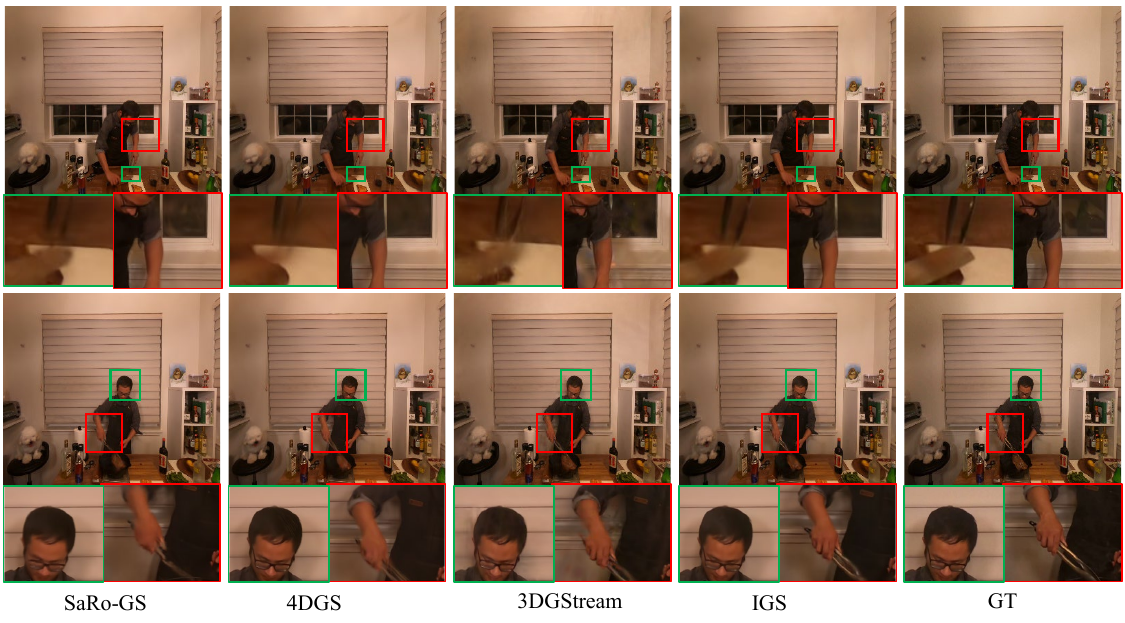}
        \vspace{-0.5cm}

  \caption{Qualitative comparison from the N3DV dataset.}
  \label{fig:n3dv_ qualitative}
  \vspace{-0.5cm}
\end{figure*}

% \noindent\textbf{Cross-domain evaluation with fine-tuning:} We conducted a cross-domain evaluation on the Meeting Room Dataset, fine-tuning our model for 5 epochs on a single sequence and leaving the other two sequences for testing. The evaluation results are presented in Tab. \ref{tab:meeting}. Our method outperforms 3DGStream in rendering quality, train time, and storage efficiency, achieving streaming with just 2.77s of per-frame reconstruction time, a significant improvement over 3DGStream. This demonstrates the effectiveness and generalizability of our approach, as it enables efficient dynamic scene modeling with streaming capabilities in new environments after only a brief fine-tuning, without requiring per-frame optimization.
% A qualitative comparison of rendering quality can be seen in Fig. \ref{fig:meeting}. Compared to 3DGStream, which produces artifacts near moving objects, our method yields more accurate motion during large displacements, resulting in improved performance in temporally complex scenes. 

\noindent\textbf{Cross-domain evaluation:} We performed a cross-domain evaluation on the Meeting Room Dataset using a model trained on N3DV. The evaluation results are presented in Tab. \ref{tab:meeting}. Our method outperforms 3DGStream in rendering quality, train time, and storage efficiency, achieving streaming with just 2.77s of per-frame reconstruction time, a significant improvement over 3DGStream. This demonstrates the effectiveness and generalizability of our approach, as it enables efficient dynamic scene modeling with streaming capabilities in new environments, without requiring per-frame optimization.
A qualitative comparison of rendering quality can be seen in Fig. \ref{fig:meeting}. Compared to 3DGStream, which produces artifacts near moving objects, our method yields more accurate motion during large displacements, resulting in improved performance in temporally complex scenes.

% \begin{table}
% \caption{Comparison on the Meeting Room dataset. $ \dagger$ indicates that the evaluation was performed using the official code in the same experimental environment as ours, including the same initial point cloud.}
% % \vspace{-10pt}
% \resizebox{0.5\textwidth}{!}{\begin{tabular}{lccccc}
%     \toprule
% \multirow{2}{*}{Method}                            & PSNR$\uparrow$                      &Latency$\downarrow$ & Training 
%  $\downarrow$ & Render$\uparrow$    & Storage$\downarrow$  \\  
%  &(dB)&(s)&(mins)&(FPS)&(MB)\\   \midrule

% 3DGStream\dagger & 28.36 & 9.74 & 0.19 & 252      & 4.0       \\
% Ours-s          & 29.35 & 0.97 & 0.05 & 252 &   1.26                \\   
% Ours-l          & 30.16 & 1.40 & 0.05 & 252 &   1.26                  \\   
% \bottomrule
% \end{tabular}}
% \label{tab:meeting}
% % \vspace{5pt}
% % \vspace{-0.6cm}
% \end{table}

\begin{table}
\centering
\caption{Comparison on the Meeting Room dataset. $ \dagger$ indicates that the evaluation was performed using the official code in the same experimental environment as ours, including the same initial point cloud.}
\vspace{-10pt}
\resizebox{0.45\textwidth}{!}{\begin{tabular}{lccccc}
    \toprule
\multirow{2}{*}{Method} & PSNR$\uparrow$    & Train 
 $\downarrow$ & Render$\uparrow$    & Storage$\downarrow$  \\  
 &(dB)&(s)&(FPS)&(MB)\\   \midrule

3DGStream\cite{sun20243dgstream}$\dagger$ & 28.36 & 11.51 & 252      & 7.59       \\
Ours-s          & 29.24  & \textbf{2.77} & 252 &   \textbf{1.26}                \\   
Ours-l          & \textbf{30.13}  & 3.20 & 252 &   \textbf{1.26}                  \\   
\bottomrule
\end{tabular}}
\label{tab:meeting}
% \vspace{5pt}
% \vspace{-0.5cm}
\end{table}

\begin{table}[h]
\centering
\caption{Ablation Study Results}
\vspace{-10pt}
\resizebox{0.45\textwidth}{!}{\begin{tabular}{lccccc}
    \toprule
\multirow{2}{*}{Method}                            & PSNR$\uparrow$                      &Train$\downarrow$      & Storage$\downarrow$  \\  
 &(dB)&(s)&(MB)\\   \midrule
No-pretrained optical flow model & 31.07 & 2.65  & 7.90       \\
No-projection-aware feature lift      & 32.95 & \textbf{2.38} &   7.90                \\   
No-points bounded refinement     & 33.23 & 3.02&   110.26           \\ 
Ours-s(full)          & \textbf{33.62} & 2.67 &   7.90                  \\ 
\bottomrule
\end{tabular}}
\label{tab:ablation}
% \vspace{5pt}
\vspace{-0.6cm}
\end{table}

\subsection{Ablation Study}
\label{sec:ablation}
\noindent\textbf{The use of the pretrained optical flow model:}
% 我们使用了预训练的Opital-flow Model来提取image pair的flow embedding,再将其Lift到3D空间。为了验证光流模型的有效性，我们进行了消融实验，将预训练的光流模型替换为没有预训练参数的4层的UNet，与整体模型一起训练。结果如表4所示，证明了我们对对2D光流先验的充分利用
% We used a pretrained optical flow model to extract flow embeddings from image pairs, which are then lifted into 3D space. To validate the effectiveness of this optical flow model, we conducted an ablation study by replacing the pretrained optical flow model with a 4-layer UNet without pretrained parameters, training it jointly with the overall model. The results, shown in Tab. \ref{tab:ablation}, demonstrate the benefit of leveraging the 2D optical flow prior in our approach.
We used a pretrained optical flow model to extract flow embeddings from image pairs, which are then lifted into 3D space. To validate its effectiveness, we replaced the pretrained model with a 4-layer UNet without pretrained parameters and trained it jointly with the overall model. The results in Tab. \ref{tab:ablation} highlight the benefit of using the 2D prior.

\noindent\textbf{Projection-aware 3D Motion Feature Lift:}
% Projection-aware 3D Motion feature lifting 
% We uses a projection-based approach to lift multi-view 2D motion feature maps into 3D space. This method allows for accurate feature acquisition by effectively relating 3D anchor points to 2D features. To evaluate the effectiveness of this module, we replaced the projection-based feature acquisition with a Transformer-based approach, applying cross-attention between image features and anchor points augmented with positional embeddings through a 4-layer Transformer block. As shown in Tab. \ref{tab:ablation}, Projection-aware Feature Lift proves crucial to the performance of IGS, with only a slight increase in train time.
We use a projection-based approach to lift multi-view 2D motion feature maps into 3D space, accurately linking 3D anchor points to 2D features. To evaluate its effectiveness, we replaced this method with a Transformer-based approach using cross-attention between image features and anchor points, enhanced with positional embeddings through a 4-layer Transformer block. As shown in Tab. \ref{tab:ablation}, Projection-aware Feature Lift is crucial for IGS performance, with only a slight increase in training time.

\begin{figure}[h]
% \vspace{-0.5cm}

  \centering
    % \fbox{\rule{0pt}{2.5in} \rule{0.9\linewidth}{0pt}}
    % \setlength{\abovecaptionskip}{0.cm}

  \includegraphics[width= \linewidth]{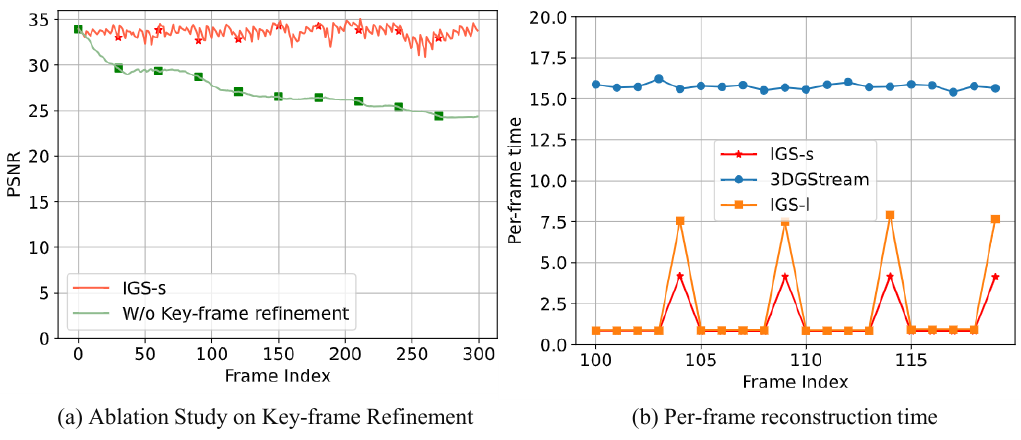}

  % \Description{ Fully described in the text.}
    \vspace{-0.4cm}

    \caption{(a)Ablation Study on Key-frame Refinement. (b)Per-frame reconstruction time.}
  \label{fig:keyframe_ab_1}

\end{figure}

\noindent\textbf{Key-frame guided Streaming:}
We employ a key-frame-guided strategy to address error accumulation in streaming and to enhance reconstruction quality. Keyframes are selected and refined through Max-points-bounded Refine. Without this refinement, AGM-Net would rely solely on Gaussians propagated from the last keyframe, resulting in accumulated errors that significantly impact performance, as shown in Fig. \ref{fig:keyframe_ab_1} (a). We also evaluate the effect of max-points bounding during refinement, as shown in Tab. \ref{tab:ablation}. Without point limits, storage requirements increase substantially, and overfitting causes a decline in view quality.
% \begin{figure}
%   \centering
%   \begin{subfigure}{0.68\linewidth}
%     \fbox{\rule{0pt}{2in} \rule{.4\linewidth}{0pt}}
%     \caption{An example of a subfigure.}
%     \label{fig:short-a}
%   \end{subfigure}
%   % \hfill
%   \begin{subfigure}{0.28\linewidth}
%     \fbox{\rule{0pt}{2in} \rule{.4\linewidth}{0pt}}
%     \caption{Another example of a subfigure.}
%     \label{fig:short-b}
%   \end{subfigure}
%   \caption{Example of a short caption, which should be centered.}
%   \label{fig:short}
% \end{figure}

\noindent\textbf{Key-frame selection:}
We conducted an ablation study on the interval \( w \) for setting keyframes, testing values of \( w=1 \), \( w=5 \), and \( w=10 \), with results shown in Tab. \ref{tab:w}.
When \( w=1 \), every frame becomes a keyframe, leading to excessive optimization that overfits Gaussians to training views, degrading test view quality and increasing training time and storage. 
% effectively making each frame a keyframe, every frame undergoes refinement following AGM-Net inference. However, due to the sparsity of viewpoints, excessive optimization leads to Gaussians overfitting to training views, decreasing quality on test views and increasing train time and Storage.
Conversely, with \( w=10 \), each keyframe drives the next 10 frames, but this distance weakens model performance, as it relies on assumptions about adjacent-frame similarity. The setting \( w=5 \) strikes the best balance across view synthesis quality, train time, and storage, and is thus our final choice.
% 我们对key-frame的设置间隔进行了ablation study,设置了 w=1,5,10 来进行对比，结果如表3所示。 当W=1时，actually相当于每一帧都是关键帧，每一帧在泛化的AGM-Net推理完后都会进行Optimization，但是由于视角的稀疏性，过分的进行Optimization反而会让Gaussian点过拟合于训练视角，反而降低了测试视角的质量，并且还会带来Latency和storage上的增加。而设置w=10时，每个关键帧需要通过泛化的AGM-Net推理后面的10帧，这么长的距离会影响的模型的效果，因为模型是基于相邻帧的相似性的先验，会导致质量下降。And w=5可以在试图合成质量，latency，Storage上均取得不错的效果，是我们最终采用的。

% 1.没有agm network
% 2.没有flow embedding
% 3. 不是anchor driven
% \subsubsection{Key-frame-guided Streaming}
% 1.没有key-frame-guided
% 2.没有max-bounded optimization

\begin{table}
\centering
\caption{The impact of different keyframe intervals \( w \).}
\vspace{-0.3cm}

\resizebox{0.4\textwidth}{!}{\begin{tabular}{lccccc}
    \toprule
Method                            & PSNR(dB)$\uparrow$                      &Train(s)$\downarrow$  
    & Storage(MB)$\downarrow$  \\  
 % &(dB)&(s)&(MB)\\   
 \midrule

w=1 & 33.55 & 6.38       & 36.0       \\
w=5          & \textbf{33.62} & \textbf{2.67} &   7.90                \\   
w=10          & 30.14 & 2.75  &   \textbf{1.26}                  \\   
\bottomrule
\end{tabular}}

\label{tab:w}
% \vspace{5pt}
\vspace{-0.5cm}
\end{table}
\section{Discussion}
% Apart from the first frame ,which may tack near 5mins to construct its Gaussians, we can get 每帧独立的reconstruction time,as shown in Fig 7. After 5 frames work-up, the econstruction time for each frame 呈现出周期的状态，0.8s for candidate frames and 4s for key frames
% We 可以通过分析每帧独立的reconsctruction time来进一步评估IGS的性能，as shown in Fig. 2.the econstruction time for each frame 呈现出周期的状态，for candidate frames it takes 0.8s and for key frames, it takes 4s and 7.5s for samll version and big version seperately, 这些都显著小于3DGStream 的16s
\subsection{Independent per-frame reconstruction time}
We further evaluate the performance of IGS by analyzing the independent per-frame reconstruction time, as shown in Fig. \ref{fig:keyframe_ab_1} (b). The reconstruction time for each frame exhibits a periodic pattern: for candidate frames, it takes 0.8s, while for key frames, it takes 4s and 7.5s for the small and large versions, respectively, which are significantly smaller than the 16s required by 3DGStream. 
\subsection{Limitation}
IGS is the first to use a generalized method for streaming dynamic scene reconstruction, but it has limitations that can be addressed in future work. As shown in Fig. \ref{fig:perframe}, our results exhibit jitter between adjacent frames, caused by the lack of temporal dependencies in the current framework. This makes the model more sensitive to noise. In contrast, 3DGStream assumes minimal motion between frames, yielding smoother results, but it fails in scenes with large motion (Fig. \ref{fig:meeting}). To reduce jitter, we plan to incorporate temporal dependencies into IGS, modeling them as a time series for more robust performance.

\section{Conclusion}
% 在本篇文章中，我们提出了IGS as a noval的以Streaming方式建模动态场景的方法，采用了泛化的解决思路，能实现1s多建模一帧，并保持甚至超越SOTA的渲染质量，同时维持较低的存储. 我们构建了一个泛化的AGM-Net, 通过将2D multi-view的Motion feature lift到3D anchor point上，以anchor来驱动Gaussian，能以一次模型推理得到相邻帧之间Gaussians的Motion. 此外，我们提出了Key-frame-guided Streaming的策略，通过选取关键帧序列并对关键帧进行refine，我们能避免误差累积，从而进一步提升渲染质量。充分的In-domain和cross-domain的实验证明了我们的模型的泛化性，将Steaming的方法的时延缩短了10倍，同时实现SOTA的渲染质量，render speed和存储开销.
In this paper, we propose IGS as a novel streaming-based method for modeling dynamic scenes. 
% By adopting a generalized approach, IGS enables frame-by-frame modeling with a per-frame reconstruction time of just over 2 seconds, while maintaining or even surpassing state-of-the-art rendering quality, all while keeping storage requirements low. 
With a generalized approach, IGS achieves frame-by-frame modeling in just over 2 seconds per frame, maintaining state-of-the-art rendering quality while keeping storage low.
We introduce a generalized AGM-Net that lifts 2D multi-view motion features to 3D anchor points, using these anchors to drive Gaussian motion. This allows the model to infer the motion of Gaussians between adjacent frames in a single step. Additionally, we propose a Key-frame-guided Streaming strategy, where key frame sequences are selected and refined to mitigate error accumulation, further enhancing rendering quality. Extensive in-domain and cross-domain experiments demonstrate the strong generalization capabilities of our model, reducing significant streaming average cost while achieving state-of-the-art rendering quality, render speed, and storage efficiency.

\section{Acknowledgments}
This work is financially supported for Outstanding Talents Training Fund in Shenzhen, this work is also financially supported by Shenzhen Science and Technology Program-Shenzhen Cultivation of Excellent Scientific and Technological Innovation Talents project(Grant No. RCJC20200714114435057), Guangdong Provincial Key Laboratory of Ultra High Definition Immersive Media Technology(Grant No. 2024B1212010006), National Natural Science Foundation of China U21B2012.

{
    \small
    \bibliographystyle{ieeenat_fullname}
    \bibliography{main}
}


% \clearpage
% \setcounter{page}{1}
% \maketitlesupplementary
\maketitlesupplementary

\renewcommand{\thesection}{\Alph{section}}
\renewcommand\thefigure{\Alph{section}\arabic{figure}}
\renewcommand\thetable{\Alph{section}\arabic{table}}
\setcounter{page}{1}
\setcounter{section}{0}
\setcounter{figure}{0}
\setcounter{table}{0}

% \section{Rationale}
% \label{sec:rationale}
% % 
% Having the supplementary compiled together with the main paper means that:
% % 
% \begin{itemize}
% \item The supplementary can back-reference sections of the main paper, for example, we can refer to \cref{sec:intro};
% \item The main paper can forward reference sub-sections within the supplementary explicitly (e.g. referring to a particular experiment); 
% \item When submitted to arXiv, the supplementary will already included at the end of the paper.
% \end{itemize}
% % 
% To split the supplementary pages from the main paper, you can use \href{https://support.apple.com/en-ca/guide/preview/prvw11793/mac#:~:text=Delete%20a%20page%20from%20a,or%20choose%20Edit%20%3E%20Delete).}{Preview (on macOS)}, \href{https://www.adobe.com/acrobat/how-to/delete-pages-from-pdf.html#:~:text=Choose%20%E2%80%9CTools%E2%80%9D%20%3E%20%E2%80%9COrganize,or%20pages%20from%20the%20file.}{Adobe Acrobat} (on all OSs), as well as \href{https://superuser.com/questions/517986/is-it-possible-to-delete-some-pages-of-a-pdf-document}{command line tools}.
\section{Overview}
With in the supplementary, we provide:
\begin{itemize}

    % \item Corrections to Tab. 2 and missing reference in our main paper in \cref{sec:correction}.
    % \item Experiments on complex outdoor environments in \cref{sec: enerf}.
    % \item Details of Key-frame-guided Streaming in \cref{sec: supp keyframe}
    \item Details of metrics calculation in \cref{sec:more metrics}.
    \item Details of experiment settings in \cref{sec:more implentation}.

    \item More ablation study in \cref{sec:more ablation}.
    \item More limitations and future work in \cref{sec:more limition}.
    \item More Results and in \cref{sec:more res}.
    \item Pseudocode and demo videos including:
    \begin{itemize}
        \item \texttt{IGS\_code.zip}
        \item \texttt{IGS-s\_testview.mp4}
        \item \texttt{IGS-l\_testview.mp4}
        \item \texttt{IGS\_freeview.mp4}
    \end{itemize}
    % \texttt{IGS\_code.zip},\texttt{IGS-s\_testview.mp4}, \texttt{IGS-l\_testview.mp4} and \texttt{IGS\_freeview.mp4}. 
    The complete code, pretrained weights, and the training dataset we constructed will be released as open-source after the review process is completed.
    % \item 伪代码和Demo视频in 'IGS.zip' 和 'IGS\_testviewpoint.mp4’ 和 'IGS freeviewpoint.mp4'. 完整的代码和预训练权重以及我们构建的训练数据将会在review结束后开源。
\end{itemize}

\section{Details of the metrics calculation}
\label{sec:more metrics}
% As mentioned in Sec.5 in the main paper, all the metrics are averaged over full 300-frame sequnce, including frame 0. Specifically:
% Storage: IGS所需要的存储包含第0帧和每一个key frame的Gaussian Primitives的存储和每一个candidate frame的存储。  每一个candidate frame由于是通过上一个key frame的Gaussians经过motion得到的，所以每一个keyframe我们只需要记录存在运动的点的mask，以及相应的du和drot.
% Train time: 与之前的方法相同, 我们报告了Train time, 这是针对一个多视频序列构建FFV所需要的时间，包括构建第0帧的Gaussian primitives的时间,使用AGM-Net得到候选帧的时间，对key frame进行refine的时间.then the total time在所有三百帧上进行平均. And 实际上这也是我们的per-frame reconstruction time.
As mentioned in Sec. 5 of the main paper, all metrics are averaged over the full 300-frame sequence, including frame 0, along with previous methods\cite{sun20243dgstream, li2022streaming}. Specifically:

\noindent\textbf{Storage:}
 The storage required for IGS includes the Gaussian primitives for frame 0 and each key frame, as well as the residuals for each candidate frame. Since each candidate frame is generated by applying motion from the previous key frame using AGM-Net, we only need to store the corresponding displacement ($du$) and rotation ($drot$), along with the mask of points with motion. We report the average storage requirements over the 300 frames.

\noindent\textbf{Train:} In line with previous methods\cite{sun20243dgstream, li2022streaming}, we report the training time, which refers to the average time required to construct an Free-Viewpoint Video from a multi-view video sequence. This includes the time for constructing the Gaussian primitives for frame 0, generating candidate frames using AGM-Net, and refining the key frames. The total time is averaged over all 300 frames, which corresponds to our per-frame reconstruction time.

\section{More implementation details}
\label{sec:more implentation}
%主要讲第一帧的实现。包括sh系数。
% 我们对每个场景下第一帧高斯点的重建效果如表x所示，我们对N3DV的场景设置了sh degree 为3，while其余两个数据集中的sh degree为1，以避免视角稀疏带来的过拟合. 在进行Max points bounded refinement时，所有场景采用同样的学习率参数，while the learning rate for position and rotation 是3DGS中的十倍，其余与3DGS的设置相同。 The Max points num N_max的设置与场景的第0帧高斯数有关，and 我们对N3DV设置Nmax = 15000,对meeting room设置Nmax=40000,对ENeRF-Outdoor设置Nmax=400000.
The reconstruction quality of Gaussian primitives for the first frame in each scenario is summarized in Tab. \ref{tab:firstframe}. For the N3DV scenes, we set the SH degree to 3, whereas for Meeting Room, it was set to 1 to mitigate overfitting caused by sparse viewpoints. During the Max Points Bounded Refinement process, all scenarios used the same learning rate settings. Specifically, the learning rate for position and rotation was set to ten times that in 3DGS, while the other parameters were kept consistent with 3DGS.

The Max Points Number \(N_{\text{max}}\) was determined based on the number of Gaussians in the initial frame of each scene. Specifically, \(N_{\text{max}}\) was set to 150,000 for N3DV, 40,000 for the Meeting Room dataset.
% 在进行Key-frame guided
% \begin{table}[h]
% \centering
% \caption{Reconstruction results of Gaussian models for the first frame in each scenario.}
% \vspace{-10pt}
% \resizebox{\linewidth}{!}{\begin{tabular}{lccccc}
%     \toprule
% \multirow{2}{*}{Scene} & PSNR$\uparrow$    & Train 
%  $\downarrow$     & Storage$\downarrow$  &Points\\  
%  &(dB)&(s)&(MB)&Num\\  
%  \midrule
%  N3DV\cite{li2022neural}\\
% \midrule
% cur roasted beef & 33.96 & 287    & 36    &149188   \\
% sear steak          & 34.03  & 287 & 35 &143996               \\   
%  \midrule
%  Meeting room\cite{li2022streaming}\\
% \midrule
% trimming & 30.36 & 540     & 3.9  & 37432   \\
% vrheadset   & 30.68  & 540 &  4    &38610    \\   
%  \midrule
%  ENeRF-Outdoor\cite{lin2022efficient}\\
% \midrule
% actor2\_3 & 26.80 & 282    & 40  &379654    \\
% actor5\_6 & 26.72 & 282 &  39    &372744   \\   
% \bottomrule
% \end{tabular}}
% \label{tab:firstframe}
% % \vspace{5pt}
% % \vspace{-0.5cm}
% \end{table}

\begin{table}[h]
\centering
\caption{Reconstruction results of Gaussian models for the first frame in each scenario.}
\vspace{-10pt}
\resizebox{\linewidth}{!}{\begin{tabular}{lccccc}
    \toprule
\multirow{2}{*}{Scene} & PSNR$\uparrow$    & Train 
 $\downarrow$     & Storage$\downarrow$  &Points\\  
 &(dB)&(s)&(MB)&Num\\  
 \midrule
 N3DV\cite{li2022neural}\\
\midrule
cur roasted beef & 33.96 & 287    & 36    &149188   \\
sear steak          & 34.03  & 287 & 35 &143996               \\   
 \midrule
 Meeting room\cite{li2022streaming}\\
\midrule
trimming & 30.36 & 540     & 3.9  & 37432   \\
vrheadset   & 30.68  & 540 &  4    &38610    \\   
%  \midrule
%  ENeRF-Outdoor\cite{lin2022efficient}\\
% \midrule
% actor2\_3 & 26.80 & 282    & 40  &379654    \\
% actor5\_6 & 26.72 & 282 &  39    &372744   \\   
\bottomrule
\end{tabular}}
\label{tab:firstframe}
% \vspace{5pt}
% \vspace{-0.5cm}
\end{table}

% 在进行Streaming Inference时，我们根据不同数据集中不同的第0帧点的个数
\section{More ablation study}
\label{sec:more ablation}
% 我们在实验中还考虑在AGM-Net中加入更多的模块，但经过实验后发现他们没有实现预期的作用，消融验证如下：
% \noindent\textbf{Attention-based view fusion:} 在进行Projection-awre Motion Feature lift时，我们考虑为不同视角的特征赋予不同的权重来结合特征，而不是像.Eq. 2中进行简单的平均. 具体而言, 我们对一个anchor在没个视角下的获取到的特征与该视角的位姿的embedding进行concatenate，then将这Nv个特征进行self-attention,then we do Softmax to get the weights用来聚合多个视角的特征. 实验结果如表所示, 加上这个模块后在N3DV的测试场景中并没有增益，这是由于N3DV中是forward facing的场景，相机视角之间带来的差异不显著，而在360°的场景中这个模块可以作为一个尝试的方向，as our future work.
% \noindent\textbf{Occlusion-aware projection:} 在进行Projection-awre Motion Feature lift时,我们尝试考虑anchor point在投影时的遮挡带来的影响. 我们尝试使用Point Rasterization and each pixel only 对应一个可见的anchor point. 实验结果如表所示， 由于我们是对anchor point进行的投影，which 相比于Gaussian稀疏很多，所以并不存在显著的遮挡，而反而采用光栅化的方式进行投影，会降低获得特征的准确性.
In our experiments, we also explored incorporating additional modules into AGM-Net. However, the results showed that these modules did not achieve the expected improvements. The ablation studies are detailed as follows:

\noindent\textbf{Attention-Based View Fusion:}  
During the Projection-Aware Motion Feature Lift, we consider assigning different weights to features from different viewpoints instead of using the simple averaging method described in Eq.2 of the main paper. Specifically, for an anchor, the features obtained from each viewpoint are concatenated with the embedding of the corresponding viewpoint's pose. These \(N_v\) features were then processed through self-attention, followed by a Softmax operation to compute the weights for aggregating the multi-view features. The experimental results, as shown in \cref{tab:ablation_supp}, indicate that adding this module doesn't yield improvements on the test scenes of N3DV. This is likely because N3DV features forward-facing scenes, where differences between camera viewpoints are not significant. However, for 360° scenes, this module could be a promising direction for future work.  

\noindent\textbf{Occlusion-Aware Projection:}  
We also attempted to account for occlusion effects during the Projection-Aware Motion Feature Lift by considering how anchor points might be obscured during projection. Specifically, we employ point rasterization\cite{ravi2020pytorch3d}, ensuring that each pixel corresponds to only one visible anchor point. The experimental results, shown in \cref{tab:ablation_supp}, reveal that this approach doesn't improve performance. Since we project anchor points, which are much sparser compared to Gaussian points, significant occlusion effects are rare. Moreover, using rasterization for projection reduces the accuracy of feature extraction.
\begin{table}[h]
\centering
\caption{More ablation study results.}
\vspace{-10pt}
\resizebox{0.45\textwidth}{!}{\begin{tabular}{lccccc}
    \toprule
Method                           & PSNR(dB)$\uparrow$                         \\  
   \midrule
Add-Attention-based view fusion      & 33.58              \\   
Add-Occulusion aware projection     & 33.50           \\ 
Ours-s          & \textbf{33.62}              \\ 
\bottomrule
\end{tabular}}
\label{tab:ablation_supp}
% \vspace{5pt}
\vspace{-0.6cm}
\end{table}

\section{Mode discussion}
\subsection{Frame Jittering}
\begin{figure}[h]
  % \vspace{-0.2cm}

  \centering
    % \fbox{\rule{0pt}{2.5in} \rule{0.9\linewidth}{0pt}}
  \includegraphics[width=\linewidth]{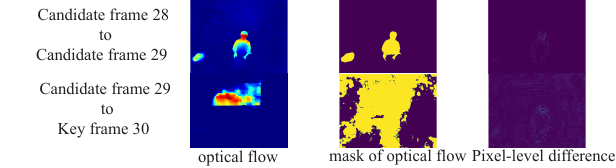}
  
  \caption{The difference between renderings of Adjacent frames}
  \label{fig:flow_compare}
  % \vspace{-0.52cm}
\end{figure}

As shown in the supplementary video, frame jittering in our method mainly occurs in static background areas. Comparing adjacent frames \cref{fig:flow_compare}, we observe that key-frame optimization causes disturbances in the background, while no such issue arises between adjacent candidate frames. This suggests that key-frame optimization deforms Gaussians in the background, particularly with floaters \cref{fig:floater}. In the moving foreground, AGM-Net prevents jitter seen in 3DGStream by smoothing point deformation. A potential solution is to segment the scene into foreground and background and apply the segmentation mask during key-frame optimization. A more robust first-frame reconstruction in sparse views could also help. 

\subsection{The impact of the number of anchor points}

\begin{figure}[h]
  % \vspace{-0.2cm}

  \centering
    % \fbox{\rule{0pt}{2.5in} \rule{0.9\linewidth}{0pt}}
  \includegraphics[width=\linewidth]{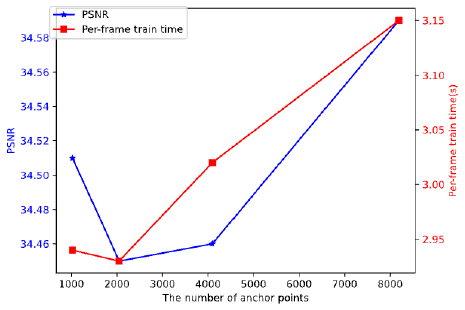}
  
  \caption{The impact of the number of anchor points}
  \label{fig:anchor_num}
  % \vspace{-0.52cm}
\end{figure}

% We tested performance with different numbers of anchor points (results shown right). The number of anchor points does not significantly impact performance. Reducing them slightly improves per-frame training speed while maintaining view synthesis quality. The main bottleneck for speed improvement is key-frame optimization (Section 6.1). To address this, we plan to build a more robust AGM-Net to reduce reliance on key-frame optimization, allowing for longer key-frame intervals and better parallelism in candidate frame processing. 
We tested performance with varying numbers of anchor points , shown in \cref{fig:anchor_num}. The number of anchor points has little impact on performance.

\section{More limitations and future work}
\label{sec:more limition}
There are additional limitations that constrain the performance of IGS, which also present opportunities for future research directions. 

\begin{figure}[h]
  \centering
    % \fbox{\rule{0pt}{2.5in} \rule{0.9\linewidth}{0pt}}
    % \setlength{\abovecaptionskip}{0.cm}

  \includegraphics[width= 
      \linewidth]{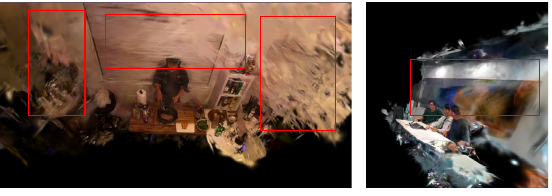}
        \vspace{-0.5cm}

  \caption{Bad Case in first-frame reconstruction:
Due to sparse viewpoints, floaters are present around moving objects, which negatively impact our streaming performance and lead to issues such as background jitter.}
  \label{fig:floater}
  % \vspace{-0.5cm}
\end{figure}

First, the performance of streaming-based dynamic scene reconstruction is influenced by the quality of static reconstruction in the first frame\cite{sun20243dgstream}. Poor reconstruction in the first frame, such as the presence of excessive floaters around moving objects as shown in \cref{fig:floater}, can degrade the performance of AGM-Net. Although addressing static reconstruction is beyond the scope of our work, adopting more robust static reconstruction methods could enhance the results of dynamic scene reconstruction.  
Second, AGM-Net has been trained on four sequences from the N3DV indoor dataset. The limited size of the training data constrains its generalization capability. Training on larger-scale multi-view video sequences is a promising direction for future improvements. Notably, our method only relies on view synthesis loss for supervision, making it easier to incorporate large-scale datasets without requiring annotated ground truth.  
Finally, our current approach injects depth and view conditions into the embeddings of an optical flow model to enable awareness of 3D scene information. Leveraging more accurate long-range optical flow\cite{wu2023accflow} or scene flow\cite{Mehl2023,teed2021raft3d} methods could further improve our results.  
% Finally, due to differences in scale and camera poses across datasets, our model currently requires fine-tuning for 5 epochs when applied to a new dataset. Incorporating data from diverse scales during training to build a scale-invariant and generalizable model could further enhance the robustness of IGS.  
% 有更多的limitations会限制IGS的性能并可以作为未来的研究方向. 首先，Streaming的动态场景重建的性能会受到第一帧的静态重建效果的影响. 如果第一帧的重建效果不佳，如在运动的物体周围有较多的floter，会影响AGM-Net的性能。虽然这不是我们工作的目的，但更鲁棒的静态重建方法会对我们的动态场景重建效果有增益. 此外，AGM-Net目前在一个室内数据集N3DV的四个序列上进行了训练，训练数据量的大小会限制它的泛化性能。在更大规模的多视角的视频序列上进行训练会作为我们后续的提升方向。并且我们的方法只需要视图合成的损失作为监督，更容易去加入更大规模的数据而不需要表注GT。第三，目前我们是对光流模型的embedding加以depth和view的condition进行finetune，以让其感知3D的场景信息，更准确的长距离光流或场景流方法可以作为我们的一个改进方向。最后，由于不同数据集的尺度和相机位姿的差异，目前我们的模型在一个新的数据集上需要在一个序列上finuetune 5个epoch，将多种尺度的数据加入训练来构建一个场景尺度free的泛化模型能进一步提升IGS的鲁棒性.
\section{More results}
\label{sec:more res}
% 在N3DV上的逐场景的与previous SOTA的对比结果如表1.Similarly,for the Plenoptic Video dataset, the per-scene
% evaluation results compared 3DGStream are shown in Table 2.

% In comparison with 3DGStream, more qualitative comparisons are presented
% in Fig. 2 and Fig. 3.

\begin{figure*}
  \centering
    % \fbox{\rule{0pt}{2.5in} \rule{0.9\linewidth}{0pt}}
    % \setlength{\abovecaptionskip}{0.cm}

  \includegraphics[width= 
      0.9\linewidth]{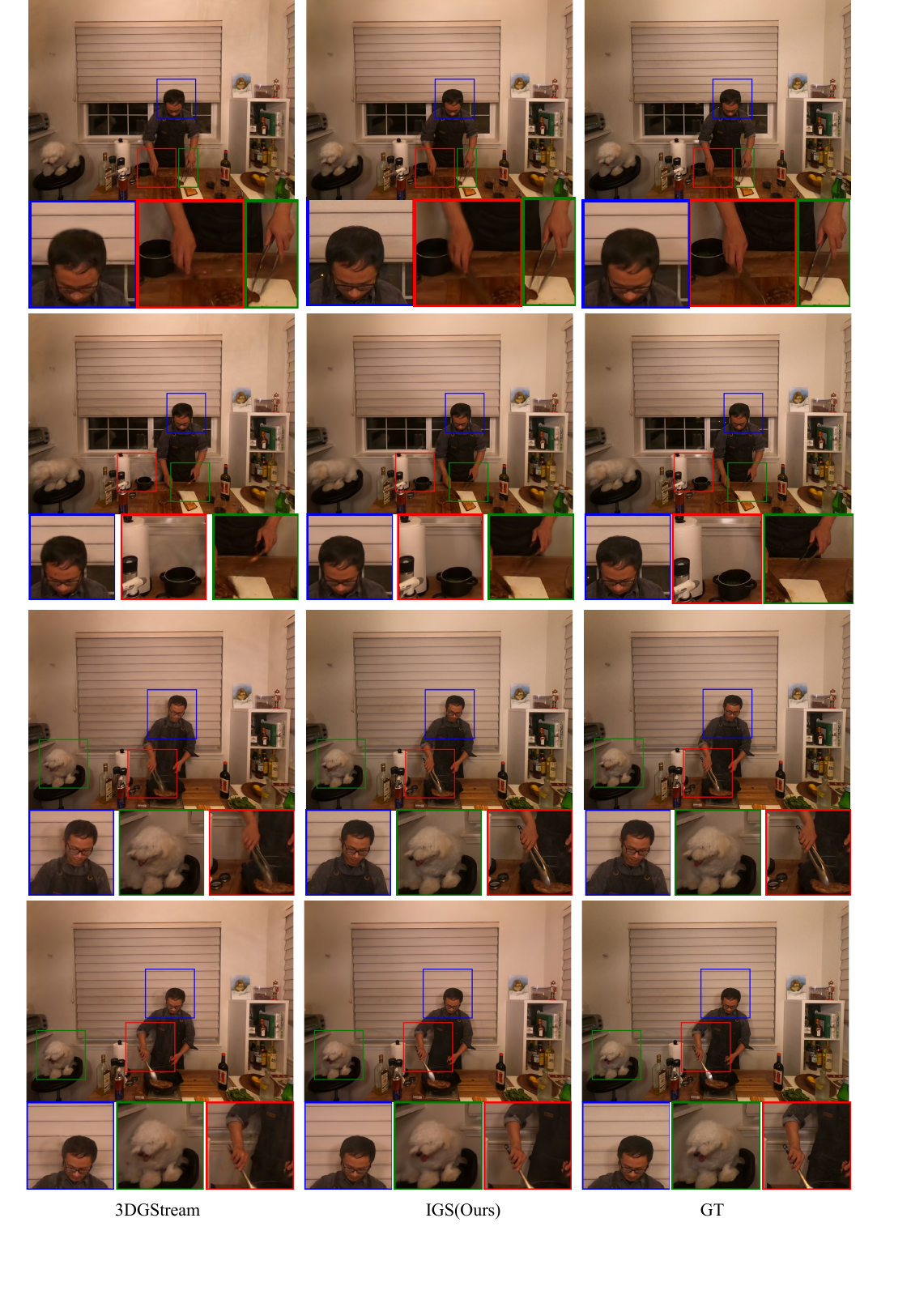}
        \vspace{-0.5cm}

  \caption{Qualitative comparison from the N3DV dataset.}
  \label{fig:supp_n3d_ qualitative}
  \vspace{-0.5cm}
\end{figure*}

% 我们提供了N3DV中的测试场景sear steak下的重建得到的test view下的视频和一个Free-Viewpoint Video using our IGS-l. The viewpoints of the Free-Viewpoint Video
% were uniformly sampled on a sphere to validate the capabilities of
% our IGS in free-viewpoint interaction with dynamic scenes. The results are shown in the video file IGS_test_view.mp4和IGS_freeviewpoint.mp4
The per-scene comparison results on the N3DV dataset against previous SOTA methods \cite{kplanes,yang2023gs4d,Wu_2024_CVPR,li2023spacetime,10.1145/3664647.3681463,li2022streaming,sun20243dgstream} are shown in \cref{tab:n3d-persene}. Further qualitative comparisons with 3DGStream\cite{sun20243dgstream} are illustrated in \cref{fig:supp_n3d_ qualitative}.
% Fig.\ref{fig:supp_n3d_ qualitative} and \ref{fig:enerf_ qualitative}.

Additionally, we provide videos showcasing the reconstruction results for the sear steak test scene from N3DV, including a Test-Viewpoint Video and a Free-Viewpoint Video generated using IGS. For the Free-Viewpoint Video, the viewpoints are uniformly sampled on a sphere to highlight the ability of our IGS to support free-viewpoint interaction with dynamic scenes. The results are available in the video files \texttt{IGS-s\_testview.mp4}, \texttt{IGS-l\_testview.mp4} and \texttt{IGS\_freeview.mp4}.

% \section{}

\begin{table}
\centering
\caption{Per-scene results on N3DV}
% \vspace{-10pt}
\resizebox{0.5\textwidth}{!}{\begin{tabular}{lcccccc}
    \toprule
\multirow{2}{*}{Method}    & \multicolumn{3}{c}{cut roasted beef} & \multicolumn{3}{c}{sear steak}\\                     
\cmidrule(r){2-4} \cmidrule(r){5-7}
~  &PSNR(dB)$\uparrow$ & DSSIM$\downarrow$ & LPIPS$\downarrow$ &PSNR(dB)$\uparrow$ & DSSIM$\downarrow$ & LPIPS$\downarrow$ \\ 
\midrule
 Offline training\\ \midrule
Kplanes\cite{kplanes}       & 31.82      & 0.017  &-  &32.52     & 0.013  & -   \\
Realtime-4DGS\cite{yang2023gs4d}   & 33.85   & -  &-      &33.51   & -  & -     \\
4DGS\cite{Wu_2024_CVPR}    & 32.90      & 0.022  &-      &32.49     & 0.022  & -   \\
Spacetime-GS\cite{li2023spacetime}     & 33.52     & \textbf{0.011}     & \textbf{0.036}   & 33.89   & \textbf{0.009}  & \textbf{0.030}           \\
Saro-GS\cite{10.1145/3664647.3681463}    & \underline{33.91}    & 0.021  & \underline{0.038}    & 33.89    &\underline{0.010} & 0.036    \\
\midrule
 Online training\\ \midrule
StreamRF\cite{li2022streaming}  & 31.81  & -  & -     & 32.36   & -  & -            \\
3DGStream\cite{sun20243dgstream} & 33.21  & -     & -      & 33.01   & -    & -               \\
3DGStream\cite{sun20243dgstream}$\dagger$ & 32.39  & \underline{0.015} &0.042   & 33.12 &0.014 &   0.036               \\
Ours-s          & 33.62 &0.012 &0.048 &   \underline{34.16}    &\underline{0.010}      &0.038      \\   
Ours-l       & \textbf{33.93}  & \textbf{0.011} &0.043&   \textbf{34.35}      &\underline{0.010}  &\underline{0.035}         \\   
\bottomrule
\end{tabular}}
\label{tab:n3d-persene}
% \vspace{5pt}
\end{table}

% \begin{figure*}[h]
%   \centering
%     % \fbox{\rule{0pt}{2.5in} \rule{0.9\linewidth}{0pt}}
%     % \setlength{\abovecaptionskip}{0.cm}

%   \includegraphics[width= 
%       \linewidth]{pic/supp_enerf.pdf}
%         \vspace{-0.5cm}

%   \caption{Qualitative comparison from the ENeRF-Outdoor dataset.}
%   \label{fig:enerf_ qualitative}
%   \vspace{-0.5cm}
% \end{figure*}

% {
%     \small
%     \bibliographystyle{ieeenat_fullname}
%     \bibliography{main}
% }

\end{document}